\documentclass{new_tlp}
\usepackage{times}
\usepackage{helvet}
\usepackage{courier}
\usepackage{amsmath}
\usepackage{algorithm}

\usepackage{amssymb}
\usepackage{graphicx}
\usepackage{color}
\usepackage{url}
\usepackage{listings}

\usepackage{wrapfig}
\newcommand{\cblu}{\color{black}}
\newcommand{\cred}{\color{black}}

\long\def\BOC#1\EOC{\message{(Commented text )}}
\long\def\BOCC#1\EOCC{\message{(Commented text )}}
\long\def\BOCCC#1\EOCCC{\message{(Commented text )}}
\long\def\optional#1{\empty}
\long\def\NB#1{\bigskip[{\cblu {\bf N.B.} #1}]\bigskip}
\long\def\NBB#1{}

\def\ar{\leftarrow}
\def\bi{\begin{itemize}}
\def\ei{\end{itemize}}
\def\beq{\begin{equation}}
\def\eeq#1{\label{#1}\end{equation}}
\def\ba{\begin{array}}
\def\ea{\end{array}}
\def\j#1{\hbox{\it #1\/}}

\def\sm{\rm SM}

\def\no{\j{not}}
\def\sneg{\sim\!\!}

\def\ar{\leftarrow}

\def\no{\j{not}}
\def\mvis{\!=\!}
\def\false{\hbox{\sc false}}
\def\true{\hbox{\sc true}}

\def\proof{\noindent{\bf Proof}.\hspace{3mm}}

\DeclareSymbolFont{AMSa}{U}{msa}{m}{n}
\DeclareMathSymbol{\square}{\mathord}{AMSa}{"03}

\def\mvis{\!=\!}
\def\mu#1{\mathit{\underline{#1}}}
\def\lpmln{{\rm LP}^{\rm{MLN}}}
\def\pbcp{p{\cal BC}+}

\def\caused{\hbox{\bf caused}}
\def\iif{\hbox{\bf if}}
\def\init{\hbox{\bf initially}}
\def\after{\hbox{\bf after}}

\def\inertial{\hbox{\bf inertial}}

\def\reward{\hbox{\bf reward}}
\def\observed{\hbox{\bf observed}}

\def\observed{\hbox{\bf observed}}

\def\:{\!:\!}

\def\bi{\begin{itemize}}
\def\ei{\end{itemize}}
\def\ii{\medskip\item}

\newtheorem{prop}{Proposition}
\newtheorem{thm}{Theorem}
\newtheorem{cor}{Corollary}
\newtheorem{definition}{Definition}
 
\newtheorem{example}{Example}

\mathchardef\mhyphen="2D 

\def\dtlpmln{{\rm DT}{\mhyphen}{\rm LP}^{\rm{MLN}}}

\begin{document}

\title{Bridging Commonsense Reasoning and Probabilistic Planning via a Probabilistic Action Language} 

\author{Yi Wang$^*$, Shiqi Zhang$^\#$, Joohyung Lee$^*$ \\
$^*$Arizona State University, USA \ \ \  $^\#$ SUNY Binghamton, USA}

\maketitle

\begin{abstract}
To be responsive to dynamically changing real-world environments, an intelligent agent needs to perform complex sequential decision-making tasks that are often guided by commonsense knowledge. The previous work on this line of research led to the framework called {\em interleaved commonsense reasoning and probabilistic planning} (i{\sc corpp}), which used P-log for representing commmonsense knowledge and Markov Decision Processes (MDPs) or Partially Observable MDPs (POMDPs) for planning under uncertainty. A main limitation of i{\sc corpp} is that its implementation requires non-trivial engineering efforts to bridge the commonsense reasoning and probabilistic planning formalisms. 
In this paper, we present a unified framework to integrate i{\sc corpp}'s reasoning and  planning components. In particular, we extend probabilistic action language $\pbcp$ to express utility, belief states, and observation as in POMDP models. Inheriting the advantages of action languages, the new action language provides an elaboration tolerant representation of POMDP that reflects commonsense knowledge. The idea led to the design of the system {\sc pbcplus2pomdp}, which compiles a $\pbcp$ action description into a POMDP model that can be directly processed by off-the-shelf POMDP solvers to compute an optimal policy of the $\pbcp$ action description.  Our experiments show that it retains the advantages of i{\sc corpp} while avoiding the manual efforts in bridging the commonsense reasoner and the probabilistic planner.  

(The article is under consideration for acceptance in TPLP.)
\end{abstract}

\section{Introduction}

Intelligent agents frequently need to perform complex sequential decision making toward achieving goals that require more than one action, in which the agent's utility depends on a sequence of decisions. A common task is to find the policy that maximizes the agent's utility when the environment is partially observable, i.e., the agent knows only partial information about the current state. 
Partially Observable Markov Decision Processes (POMDPs) \cite{kaelbling1998planning} have been widely used for that purpose. It assumes partial observability of underlying states and can model nondeterministic state transitions and local, unreliable observations using probabilities, and plan toward maximizing long-term rewards under such uncertainties.
However, as a very general mathematical framework, POMDPs are not equipped with built-in constructs for representing commonsense knowledge. 

Recent works \cite{zhang15corpp,zhang15mixed} aim at embracing commonsense knowledge into probabilistic planning. 
In that line of research, a reasoner was used for state estimation with contextual knowledge, and a planner focuses on selecting actions to maximize the long-term reward.  More recently, probabilistic logical knowledge has been used for reasoning about both the current state and the dynamics of the world, resulting in the framework called i{\sc corpp}~\cite{zhang17dynamically}. 
i{\sc corpp} builds on two formalisms: P-log~\cite{baral09probabilistic}  for commonsense reasoning and 
 POMDP~\cite{kaelbling1998planning} for probabilistic planning .
Reflecting the commonsense knowledge, i{\sc corpp} significantly reduces the complexity of POMDP planning while enabling robot behaviors to adapt to exogenous changes. One example domain in \cite{zhang17dynamically} demonstrates that the MDP constructed by i{\sc corpp} includes only 60 states whereas the naive way of enumerating all combinations of attribute values produces more than $2^{69}$ states.

Despite the advantages, i{\sc corpp} has the limitation that practitioners must spend non-trivial engineering efforts to bridge the gap between P-log and POMDP in its implementations. One reason is that P-log does not have the built-in notions of utility and partially observable states as in POMDP models. Thus, the work on i{\sc corpp}  acquired the transitions and their probabilities by running a P-log solver, but then the user has to manually add the information about the rewards and the belief states~\cite{zhang17dynamically}.

In this paper, we present a more principled way to integrate the commonsense reasoning and probabilistic planning components in the i{\sc corpp} framework, which serves as the main contribution of this paper.
We achieve this by extending probabilistic action language $\pbcp$ \cite{lee18aprobabilistic,wang19elaboration} to support the representation of and reasoning with utility, belief states, and observation as in POMDP models. Inheriting the advantages of action languages, the new action language provides an elaboration tolerant representation of POMDP that is convenient to encode commonsense knowledge and completely shield users from the syntax or algorithms of POMDPs. 

The second contribution is on the design of the system {\sc pbcplus2pomdp}, which can dynamically construct POMDP models given an action description in $\pbcp$, and compute action policies using off-the-shelf POMDP solvers. 
Unlike i{\sc corpp}, the semantics of $\pbcp$ and its reasoning system together support the direct generation of planning models, which can be further used for computing action policies using POMDP solvers. 
Experimental results 
show that the extended $\pbcp$ (and its supporting system) retains the advantages of i{\sc corpp} while successfully avoiding the manual efforts in bridging the gap between i{\sc corpp}'s  commonsense reasoning and probabilistic planning components. 






The paper is organized as follows. After reviewing $\pbcp$ and POMDP in Section~\ref{sec:prelim}, we extend $\pbcp$ and show how it can be used to represent POMDP models in Section~\ref{sec:pomdp-pbcp}. In Section~\ref{sec:elab}, we show how we can dynamically generate POMDP models by exploiting the elaboration tolerant representation of $\pbcp$.
We present the system {\sc pbcplus2pomdp} in Section~\ref{sec:system} and experimental results with the system in Section~\ref{sec:evaluation}. After discussing the related work in Section~\ref{sec:related}, we conclude in Section~\ref{sec:conclusion}.

\section{Preliminaries} \label{sec:prelim}

Due to the space limit, the review is brief. For more detailed reviews, we refer the reader to  \cite{lee18aprobabilistic,wang19elaboration}, or the supplementary material corresponding 
to this paper at the TPLP archives.

\subsection{Review: $\pbcp$ with Utility}  \label{ssec:pbcp}

We review $\pbcp$ as presented in~\cite{wang19elaboration}, which extends the language in~\cite{lee18aprobabilistic} by incorporating the concept of utility.

Like its predecessors ${\cal BC}$ \cite{lee13action} and ${\cal BC}$+ \cite{babb15action1}, language $\pbcp$ assumes that a propositional signature $\sigma$ is constructed from ``constants'' and their ``values.'' 
A {\em constant} $c$ is a symbol that is associated with a finite set $\j{Dom}(c)$, called the {\em domain}. 
The signature $\sigma$ is constructed from a finite set of constants, consisting of atoms $c\!=\!v$
for every constant $c$ and every element $v$ in $\j{Dom}(c)$.
If the domain of~$c$ is $\{\false,\true\}$, then we say that~$c$ is {\em Boolean}, and abbreviate $c\mvis\true$ as $c$ and $c\mvis\false$ as~$\sneg c$. 

\begin{figure}[t]
\centering
\includegraphics[width=1\textwidth]{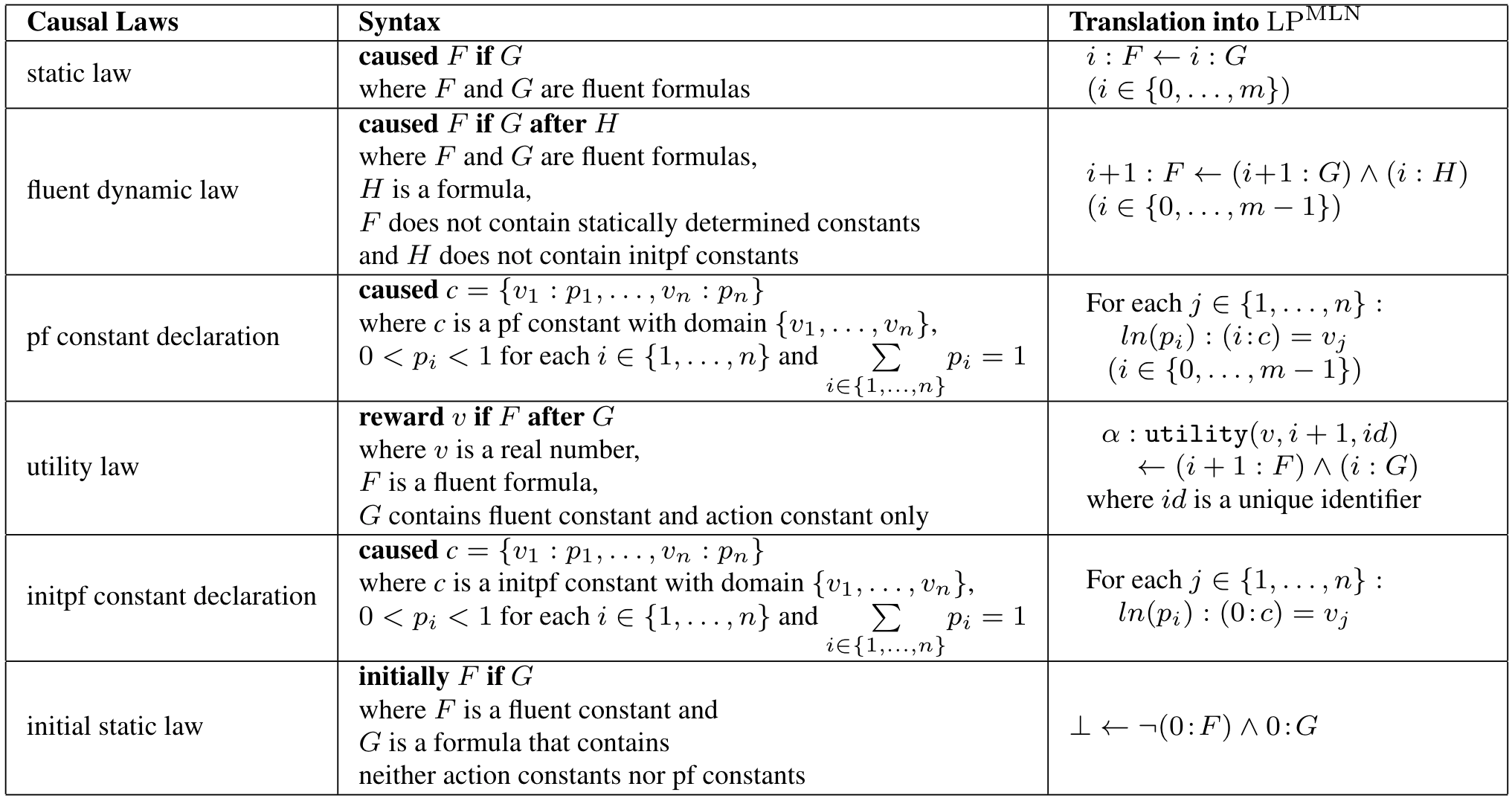}
\caption{Causal laws in $\pbcp$ and their translations into $\lpmln$}
\label{fig:pbcplus-causal-laws}
\end{figure}

There are four types of constants in $\pbcp$: {\em fluent constants}, {\em action constants},  {\em pf (probability fact) constants} and {\em  initpf (initial probability fact) constants}. Fluent constants are further divided into {\em regular} and {\em statically determined}. The domain of every action constant is restricted to Boolean. An {\em action description} is a finite set of {\em causal laws}, which describes how fluents depend on each other statically and how their values change from one time step to another. Fig.~\ref{fig:pbcplus-causal-laws} lists causal laws in $\pbcp$ and their translations into $\lpmln$ \cite{lee16weighted}.
A {\em fluent formula} is a formula such that all constants occurring in it are fluent constants. 

We use $\sigma^{fl}$ ($\sigma^{act}$, $\sigma^{pf}$, and $\sigma^{initpf}$, respectively)  to denote the set of all atoms $c\mvis v$ where $c$ is a fluent constant (action constant, pf constant, initpf constant, respectively) of $\sigma$ and $v$ is in $\j{Dom}(c)$. 
{\cred For any maximum time step $m$}, any subset $\sigma'$ of $\sigma$ and any $i\in\{0, \dots, m\}$, we use $i\!:\!\sigma^\prime$ to denote the set
$\{i\!:\!A \mid A\in\sigma^\prime\}$. For any formula $F$ of signature $\sigma$, by $i\!:\!F$ we denote the result of inserting $i\!:$ in front of every occurrence of every constant in~$F$. 

The semantics of a $\pbcp$ action description $D$ is defined by a translation into an $\lpmln$ program $Tr(D, m) = D_{init}\cup D_m$. Below we describe the essential part of the translation that turns a $\pbcp$ description into an $\lpmln$ program. 

The signature $\sigma_m$ of $D_m$ consists of  atoms of the form $i\!:\!c=v$ such that
\begin{itemize}
\item for each fluent constant $c$ of $D$, $i\in\{0, \dots, m\}$ and $v\in Dom(c)$,
\item for each action constant or pf constant $c$ of $D$, $i\in\{0, \dots, m-1\}$ and $v\in Dom(c)$.
\end{itemize}
and atoms of the form ${\tt utility}(v, i, id)$ introduced by each utility law as described in Fig.~\ref{fig:pbcplus-causal-laws}.

$D_m$ contains $\lpmln$ rules obtained from static laws, fluent dynamic laws,  utility laws, and pf constant declarations as described in the third column of Fig. \ref{fig:pbcplus-causal-laws}, as well as   $\{0\!:\!c=v\}^{\rm ch}$ for every regular fluent constant $c$ and every $v\in Dom(c)$, and
$\{i\!:\!c=\true\}^{\rm ch}, \{i\!:\!c=\false\}^{\rm ch}$ ($i\in\{0,\dots,m\!-\!1$)
for every action constant $c$ to state that the fluents at time $0$ and the actions at each time are exogenous.\footnote{$\{A\}^{\rm ch}$ denotes the choice rule $A\leftarrow \no\ \no\ A$.}
$D_{init}$ contains $\lpmln$ rules obtained from initial static laws and initpf constant declarations as described in the third column of Fig.~\ref{fig:pbcplus-causal-laws}.
Both $D_m$ and $D_{init}$ also contain constraints asserting that each constant is mapped to exactly one value in its domain. We identify an interpretation of $\sigma_m$ (or $\sigma$) that satisfies these constraints with the value assignment function mapping each constant to its value.




%
For any $\lpmln$ program $\Pi$ of signature $\sigma_1$ and any interpretation $I$ of a subset $\sigma_2$ of~$\sigma_1$, we say $I$ is a {\em residual (probabilistic) stable model} of $\Pi$ if there exists an interpretation $J$ of $\sigma_1\setminus \sigma_2$ such that $I\cup J$ is a (probabilistic) stable model of $\Pi$.

For any interpretation $I$ of $\sigma$, by $i\!:\!I$ we denote the interpretation of  $i\!:\!\sigma$ such that $i\!:\!I\models (i\!:\!c)=v$ iff $I\models c=v$. 
For $x\in \{act, fl, pf\}$, we use $\sigma^{x}_m$ to denote the subset of $\sigma_m$, which is 
$
\{i\!:\!c=v\in\sigma_m \mid\ \text{$c=v\in\sigma^{x}$}\}.
$

A {\em state} of $D$ is an interpretation $I^{fl}$ of $\sigma^{fl}$ such that $0\!:\!I^{fl}$ is a residual (probabilistic) stable model of $D_0$. A {\em transition} of $D$ is a triple $\langle s, e, s^\prime\rangle$  where $s$ and $s^\prime$ are interpretations of $\sigma^{fl}$ and $e$ is an interpretation of $\sigma^{act}$ such that $0\!:\!s \cup 0\!:\!e \cup 1:s^\prime$ is a residual stable model of $D_1$. A {\em pf-transition} of $D$ is a pair $(\langle s, e, s^\prime\rangle, pf)$, where $pf$ is a value assignment to $\sigma^{pf}$ such that $0\!:\!s\cup 0\!:\!e \cup 1:s^\prime \cup 0\!:\!pf$ is a stable model of $D_1$.


The following simplifying assumptions are made on action descriptions in $\pbcp$.
%
\begin{enumerate}
\item {\bf No concurrent execution of actions}: For all transitions $\langle s, e, s'\rangle$, we have $e\models a\!=\!\true$ for at most one action constant $a$; 
%
\item {\bf Nondeterministic transitions are determined by pf constants}: For any state $s$, any value assignment $e$ of $\sigma^{act}$, and any value assignment $pf$ of $\sigma^{pf}$, there exists exactly one state $s^\prime$ such that $(\langle s, e, s^\prime\rangle, pf)$ is a pf-transition;
\item {\bf Nondeterminism on initial states are determined by initpf constants}: For any value assignment $pf_{init}$ of $\sigma^{initpf}$, there {exists exactly one value assignment $fl$ of $\sigma^{fl}$ such that $0\!:\!pf_{init}\cup 0\!:\!fl$ is a stable model of $D_{init}\cup D_0$.}


\end{enumerate}

With the above three assumptions, the probability of a history, i.e., a sequence of states and actions, can be computed as the product of the probabilities of all the transitions that the history is composed of,  multiplied by the probability of the initial state. 

\NBB{$v$ should be a string in utility atom?}


%

\BOC
The {\em utility} of an interpretation $I$ under $\Pi$ is defined as
\[
U_{\Pi}(I) = \underset{{\tt utility}(u, {\bf t})\in I}{\sum} u .
\]
The {\em expected utility} of a proposition $A$ is defined as
\begin{equation}\label{eq:expected-utility}
E[U_{\Pi}(A)] = \underset{I\models A}{\sum}\  
    U_{\Pi}(I) \times P_{\Pi}(I\mid A) . 
\end{equation}
\EOC

A $\pbcp$ action description defines a probabilistic transition system as follows: A {\em probabilistic transition system} $T(D)$ represented by a probabilistic action description $D$ is a labeled directed graph such that the vertices are the states of $D$, and the edges are obtained from the transitions of $D$: for every transition $\langle s, e, s^\prime\rangle$  of $D$, an edge labeled $e: p, u$ goes from $s$ to $s^\prime$, where $p=P_{D_1}(1\!:\!s^\prime \mid 0\!:\!s\land 0\!:\!e)$ and $u=E[U_{D_1}(0\!:\!s\land 0\!:\!e\land 1\!:\!s')]$. \footnote{
The {\em utility} of an interpretation $I$ under $\dtlpmln$ program $\Pi$ \cite{wang19elaboration} is defined as
$
U_{\Pi}(I) = \Sigma_{{\tt utility}(u, {\bf t})\in I}\ u 
$
and the {\em expected utility} of a proposition $A$ is defined as
$ 
E[U_{\Pi}(A)] = \underset{I\models A}{\sum}\  
    U_{\Pi}(I) \times P_{\Pi}(I\mid A) 
$.
}
The number $p$ is called the {\em transition probability} of $\langle s, e ,s^\prime\rangle$, denoted by $p(s, e ,s^\prime)$, and the number $u$ is called the {\em transition reward} of $\langle s, e ,s^\prime\rangle$, denoted by $u(s, e ,s^\prime)$. 
The notion of a probabilistic transition system is essentially the same as that of a Markov Decision Process.


\subsection{Review: POMDP}  \label{ssec:pomdp}
A Partially Observable Markov Decision Processes (POMDP) is defined as a tuple
\[
  \langle S, A, T, R, \Omega, O, \gamma\rangle
\]
where 
(i) $S$ is a set of states;
(ii) 
$A$ is a set of actions;
(iii) $T: S\times A\times S\rightarrow [0, 1]$ are transition probabilities;
(vi) $R: S\times A\times S\rightarrow \mathbb{R}$ are rewards;
(v) $\Omega$ is a set of observations;
(vi) $O: S\times A\times \Omega \rightarrow [0, 1]$ are observation probabilities;
(vii) $\gamma\in [0,1]$ is a discount factor. 


A {\em belief state} is  a probability distribution over $S$.
Given the current belief state $b$, after taking action $a\in A$ and observing $o\in \Omega$, the updated belief state $b'$ can be computed as $$
    b'(s')=\eta\cdot O(o\mid s',a)\sum _{{s\in S}}T(s'\mid s,a)b(s)
$$
where $s, s'\in S$ are the current and next states respectively; $b(s)$ is the belief probability in $b$ corresponding to $s$; $b'(s')$ is the belief probability in $b'$ corresponding to $s'$; and $\eta$ is a normalizer. 

A {\em policy} $\pi$ is a function from the set of belief states to the set of actions. 
The {\em expected total reward} of a stationary policy $\pi$ starting from the initial belief state $b_0$ is
\[
\ba{rll}
V^\pi(b_0) 
& = \ \sum_{t=0}^\infty \gamma^t E\Bigl[ R(s_t,\pi(b_t), s_{t+1}) \mid b_0 \Bigr]
\ea
\]
where $b_t$ and $s_t$ are the belief state and  the state at time $t$, respectively.
The optimal policy $\pi^*$ is obtained by optimizing the long-term reward:
$
\pi ^{*}={\underset  {\pi }{{\mbox{argmax}}}}\ V^{\pi }(b_{0}).
$

\section{Representing POMDP by Extended $\pbcp$} \label{sec:pomdp-pbcp}

To be able to express partially observable states, we extend $p\cal{BC}+$ by introducing a new type of constants, called {\em observation constants}, and a new kind of causal laws called {\em observation dynamic laws}. An {\em observation dynamic law} is of the form
\begin{equation}\label{eq:observation-dynamic-law}
{\observed}\ F\ {\bf if}\ G\ \after\ H
\end{equation}
where $F$ is a formula containing {\cred no constants other than} observation constants, $G$ is a formula containing {\cred no constants other than} fluent constants, and $H$ is a formula containing no constants other than action constants and pf constants. 
\BOCC
An observation static law is of the form
\begin{equation}\label{eq:observation-static-law}
{\bf observe}\ F\ {\bf if}\ G
\end{equation}
where $F$ and $G$ contains observation constants only. 
\EOCC
Observation constants can occur only in observation dynamic laws. An observation dynamic law $r$ of the form \eqref{eq:observation-dynamic-law} is translated into the following $\lpmln$ rule:
\begin{align}
\nonumber \alpha\ &:\ (i+1\!:\!F) \leftarrow (i+1\!:\!G)\land (i\!:\!H).
\end{align}
\BOCC
and an observation static law $r$ of the form \eqref{eq:observation-static-law} is translated into the following $\lpmln$ rule:
\begin{align}
\nonumber \alpha\ &:\ (i\!:\!F) \leftarrow (i\!:\!G)
\end{align}
\EOCC

For each observation constant $obs$, 
$Dom(obs)$ contains a special value ${\tt NA}$ (``Not Applicable''). For each observation constant $obs$ in $\sigma^{obs}$ and $v\in Dom(obs)$, we include the following $\lpmln$ rule in $D_m$ to indicate that the initial value of each observation constant is exogenous:
\[
\nonumber \alpha\ :\ \{0:obs\mvis v\}^{\rm ch}
\]
and include the following $\lpmln$ rule in $D_m$ to indicate that the default value of  $obs$ is ${\tt NA}$:
\[
 \alpha\ :\ \{i:obs\mvis {\tt NA}\}^{\rm ch}  \ \ \  (i\in\{1,\dots,m\}).
\]

\NBB{This is not always the case}

For a more flexible representation, we introduce the $\iif$ clause in the pf constant declarations as
\begin{equation}\label{eq:pf-constant-extended}
\caused\ c=\{v_1:p_1, \dots, v_n:p_n\}\ \iif\ F
\end{equation}
where $c$ is a pf constant with the domain $\{v_1, \dots, v_n\}$, $0<p_i<1$ for each $i\in\{1, \dots, n\}$, $\underset{i\in\{1, \dots, n\}}{\sum}p_i=1$ and $F$ contains rigid constants only.\footnote{A {\em rigid} constant is a statically determined fluent constant for which the value is assumed not to change over time \cite{giu04}.} A pf constant declaration~\eqref{eq:pf-constant-extended} is translated into $\lpmln$ rules
\begin{equation}\label{eq:nonground-pf-atom}
  ln(p_i) : (i:c)=v_j \leftarrow F
\end{equation}
for $j\in\{0, \dots, m\}$.
In addition to Assumptions 1--3 above, we add the following assumption:
\begin{enumerate}
\item[4.] 
{\bf Rigid constants take the same value over all stable models}: for any rigid constant $c$, there exists $v\in Dom(c)$ such that $I\vDash c=v$ for all stable model $I$ of $D_m$.
\end{enumerate}
Under this assumption, the body $F$ in~\eqref{eq:nonground-pf-atom} evaluates to either $\true$ or $\false$ for all stable models of $D_m$, meaning that either \eqref{eq:nonground-pf-atom} can be removed from $D_m$, or $F$ can be removed from the body of \eqref{eq:nonground-pf-atom}. Thus, this is not an essential extension but helps us use different probability distributions by changing the condition $F$.



\BOC
{\cred 
For more flexible representations, we introduce a new type of constant called {\em rigid constant}, which intuitively represent fluents whose values do not change over time steps. A {\em rigid static law} is an expression of the form
\begin{equation}\label{eq:rigid-static-law}
\caused\ F\ \iif\ G
\end{equation}
where $F$ and $G$ contain rigid constants only. A rigid static law~\eqref{eq:rigid-static-law} is translated into $\lpmln$ rule
\[
\alpha: F\leftarrow G
\]
in $D_m$ (?).
We then extend pf constant declaration as
\begin{equation}\label{eq:pf-constant-extended}
\caused\ c=\{v_1:p_1, \dots, v_n:p_n\}\ \iif\ F
\end{equation}
where $c$ is a pf constant with domain $\{v_1, \dots, v_n\}$, $0<p_i<1$ for each $i\in\{1, \dots, n\}$, $\underset{i\in\{1, \dots, n\}}{\sum}p_i=1$ and $F$ contains rigid constants only. A pf constant declaration~\eqref{eq:pf-constant-extended} is translated into $\lpmln$ rules
\begin{equation}\label{eq:nonground-pf-atom}
ln(p_i) : (i:c)=v_j \leftarrow F
\end{equation}
for $j\in\{0, \dots, m-1\}$.
}
\EOC

\BOC
{\cred 
We use $\sigma^{obs}$ to denote the set of observation fluents. The signature $\sigma_m$ of $D_m$ is now extended with $\sigma^{obs}_m$.
}
\EOC
 
Given a $\pbcp$ action description $D$, we use ${\bf S}$ to denote the set of states, i.e, the set of interpretations $I^{fl}$ of $\sigma^{fl}$ such that $0\!:\!I^{fl}$ is a residual (probabilistic) stable model of $D_0$. We use ${\bf A}$ to denote the set of interpretations $I^{act}$ of $\sigma^{act}$ such that $0\!:\!I^{act}$ is a residual (probabilistic) stable model of $D_1$. Since we assume that at most one action is executed each time step, each element in ${\bf A}$ makes either only one action 
or none to be true. 

\begin{definition}\label{def:pbcp-pomdp}
A $p\cal{BC}+$ action description $D$, together with a discount factor $\gamma$, defines a POMDP $M(D)$
$
\langle S, A, P, R, \Omega, O, \gamma\rangle
$
where
\begin{itemize}
\item the state set $S$ is the same as {\bf S} and the action set $A$ is the same as ${\bf A}$;
\item the transition probability $P$ is defined as $P(s, a, s')= P_{D_1}(1\!:\!s'\mid 0\!:\!s, 0\!:\!a)$;
\item the reward function $R$ is defined as $R(s, a, s') = E[U_{D_1}(0\!:\!s, 0\!:\!a, 1\!:\!s')]$;
\item the observation set $\Omega$ is the set of interpretations $o$ on $\sigma^{obs}$ such that $0\!:\!o$ is a residual stable model of $D_0$;
\item the observation probability $O$ is defined as $O(s, a, o ) = P_{D_1}(1\!:\!o\mid 1\!:\!s, 0\!:\!a)$.
\end{itemize}
\end{definition}

\NBB{Initial belief state? $b_0(s) = P_{Dinit}(s)$} 

\BOCC
\begin{example}\label{eg:2-tiger}
({\bf Two Tigers Example}). Consider a variant of the well-known tiger example extended with two tigers. Each of the three doors has either a tiger or a prize behind. 
The agent can open either of the three doors. The agent can also listen to get a better idea of where the tiger is. Listening yields the correct information about where each of the two tigers is with probability $0.85$. 
This example can be represented in the extended $p\cal{BC}+$ as follows:

\hrule
\begin{tabbing}
Notation:  $l, l_1, l_2, l_3$ range over ${\tt Left}$, ${\tt Middle}$, ${\tt Right}$, $y$ ranges over ${\tt Tiger1}$, ${\tt Tiger2}$\\
Observation constants:         \hskip 4cm  \=Domains:\\
$\;\;\;$ $\j{TigerPositionObserved}(y)$                 \>$\;\;\;$ $\{{\tt Left}, {\tt Middle}, {\tt Right}, {\tt NA}\}$\\ 
Regular fluent constants:          \hskip 4cm  \=Domains:\\
$\;\;\;$ $\j{TigerPosition}(y)$                 \>$\;\;\;$ $\{{\tt Left}, {\tt Middle}, {\tt Right}\}$\\ 
Action constants:                          \>Domains:\\
$\;\;\;$ $\j{Listen}$  \>$\;\;\;$ Boolean\\
$\;\;\;$ $\j{OpenDoor}(l)$  \>$\;\;\;$ Boolean\\ 
Pf constants:                          \>Domains:\\
$\;\;\;$ $\j{Pf\_Listen}$                    \>$\;\;\;$ Boolean\\
$\;\;\;$ $\j{Pf\_FailedListen}(y)$                    \>$\;\;\;$ $\{{\tt Left}, {\tt Middle}, {\tt Right}\}$
\end{tabbing}
\hrule

A reward of $10$ is obtained for opening the door with no tiger behind.
\begin{align}
\nonumber &{\reward}\ 10\ \iif\ \j{TigerPosition}({\tt Tiger1})\mvis l_1\land \j{TigerPosition}({\tt Tiger2})\mvis l_2\ \after\ \j{OpenDoor}(l_3)\\
\nonumber & \ \ \ \ \ \ \ (l_1 \neq l_3,  l_2\neq l_3).
\end{align}
A penalty of $100$ is imposed on opening a door with a tiger behind.
\begin{align}
\nonumber &{\reward}\ -\!\!100\ \iif\ \j{TigerPosition}(y)\mvis l\ \after\ \j{OpenDoor}(l).
\end{align}
Executing the action $\j{Listen}$ has a small penalty of $1$.
\begin{align}
\nonumber &{\reward}\ -\!\!1\ \iif\ \top\ \after\ \j{Listen}.
\end{align}
Two tigers cannot be in the same position.
\begin{align}
\nonumber &{\caused}\ \bot\ \iif\ \j{TigerPosition}({\tt Tiger1})\mvis l\land \j{TigerPosition}({\tt Tiger2})\mvis l.
\end{align}
Successful listening reveals the positions of the two tigers.
\begin{align}
\nonumber &{\observed}\ \j{TigerPositionObserved}(y)\mvis l\ {\bf if}\ \j{TigerPosition}(y)\mvis l\ {\bf after}\ \j{Listen}\land \j{Pf\_Listen}.
\end{align}
Failed listening yields a random position for each tiger.
\begin{align}
\nonumber &\caused\ \j{Pf\_FailedListen}(y)=\{{\tt Left}: \frac{1}{3}, {\tt Middke}: \frac{1}{3}, {\tt Right}: \frac{1}{3}\},\\
\nonumber &{\observed}\ \j{TigerPositionObserved}(y)\mvis l\ {\bf if}\ \top\ {\bf after}\ \j{Listen}\land \sim\j{Pf\_Listen}\land \j{Pf\_FailedListen(y)} \mvis  l.
\end{align}
The positions of tigers observe the commonsense law of inertia.
\begin{align}
\nonumber &\inertial\ \j{TigerPosition}(y).
\end{align}
The action $\j{Listen}$ has a success rate of $0.85$.
\begin{align}
\nonumber &\caused\ \j{Pf\_Listen}=\{\true: 0.85, \false: 0.15\}.
\end{align}

\end{example}

\EOCC 


\section{Elaboration Tolerant Representation of POMDP} \label{sec:elab}

We illustrate the features of the extended $\pbcp$ using the ``dialog management'' example from~\cite{zhang17dynamically}, where a robot is responsible for delivering an item $i$ to person $p$ in room $r$. The robot needs to ask questions to figure out what $i$, $p$, $r$ are. The challenge comes from the robot's imperfect speech recognition capability. As a result, repeating questions is sometimes necessary. We use POMDP to model the unreliability from speech recognition, and the robot uses observations to maintain a belief state in the form of a probability distribution. 
There are two types of questions that the robot can ask:
\begin{itemize}
    \item Which-Questions: questions about which item/person/room it is, for example, ``which item is it?''
    \item Confirmation-Questions: questions to confirm whether a(n) item/person/room is the requested one, for example, ``is the requested item coffee?''
\end{itemize}
Each of the question-asking actions has a small cost. 
The robot can execute a $\j{Deliver}$ action, which consists of an item $i'$, person $p'$ and room $r'$ as arguments. A $\j{Deliver}$ action deterministically leads to the terminal state. 
A reward is obtained with $\j{Deliver}$ action, determined by to what extent $i'$, $p'$ and $r'$ matches $i$, $p$ and $r$.
For instance, when all three entries are correctly identified in the $\j{Deliver}$ action, the agent receives a large reward; when none is correctly identified, the agent receives a large penalty (in the form of a negative reward). 
Therefore, the agent has the motivation of computing action policies to minimize the cost of its question-asking actions, while maximizing the expected reward by tasking the ``correct'' delivery action. 

This example can be represented in $\pbcp$ as follows. We assume a small domain where $\j{Item} = \{\j{Coffee}, \j{Coke}, \j{Cookies}, \j{Burger}\}$, $\j{Person} = \{\j{Alice}, \j{Bob}, \j{Carol}\}$, $\j{Room} = \{R_1, R_2, R_3\}$.

\medskip\hrule
\begin{tabbing}
Notation:  $i$, $i'$ range over $\j{Item}$, $p, p'$ ranges over $\j{Person}$, $r, r'$ ranges over $\j{Room}$, $c$ ranges over $\{{\tt Yes}, {\tt No}\}$\\
Observation constant:         \hskip 6.5cm  \=Domains:\\
$\;\;\;$ $\j{ItemObs}$                 \>$\;\;\;$ $\j{Item}\cup\{{\tt NA}\}$\\ 
$\;\;\;$ $\j{PersonObs}$                 \>$\;\;\;$ $\j{Person}\cup\{{\tt NA}\}$\\ 
$\;\;\;$ $\j{RoomObs}$                 \>$\;\;\;$ $\j{Room}\cup\{{\tt NA}\}$\\ 
$\;\;\;$ $\j{Confirmed}$                 \>$\;\;\;$ $\{{\tt Yes}, {\tt No}, {\tt NA}\}$  \\ 

Regular fluent constants:          \hskip 6.2cm  \=Domains:\\
$\;\;\;$ $\j{ItemReq}$                 \>$\;\;\;$ $\j{Item}$\\ 
$\;\;\;$ $\j{PersonReq}$                 \>$\;\;\;$ $\j{Person}$\\
$\;\;\;$ $\j{RoomReq}$                 \>$\;\;\;$ $\j{Room}$\\
$\;\;\;$ $\j{Terminated}$                 \>$\;\;\;$ Boolean\\
Action constants:                          \>Domains:\\
$\;\;\;$ $\j{WhichItem}$,\  
 $\j{WhichPerson}$, \  
 $\j{WhichRoom}$, \\
$\;\;\;$ $\j{ConfirmItem}(i)$, \ 
 $\j{ConfirmPerson}(p)$, \  $\j{ConfirmRoom}(r)$, \\ 
\  
 $\;\;\;$   $\j{Deliver}(i, p, r)$  \>$\;\;\;$ Boolean\\ 
Pf constants:                          \>Domains:\\
$\;\;\;$ $\j{Pf\_WhichItem}(i)$                    \>$\;\;\;$ $\j{Item}$ \\
$\;\;\;$ $\j{Pf\_WhichPerson}(p)$                    \>$\;\;\;$ $\j{Person}$ \\
$\;\;\;$ $\j{Pf\_WhichRoom}(r)$                    \>$\;\;\;$ $\j{Room}$ \\
$\;\;\;$ $\j{Pf\_ConfirmWhenCorrect}$,\  $\j{Pf\_ConfirmWhenIncorrect}$                   \>$\;\;\;$ $\{{\tt Yes}, {\tt No}\}$ 
\end{tabbing}
\hrule
\bigskip


The action $\j{Deliver}$ causes the entering of the terminal state:
\begin{align}
\nonumber &\caused\ \j{Terminated}\ \iif\ \top \after\ \j{Deliver}(i, p, r).
\end{align}
The execution of $\j{Deliver}$ action with the room, the person and the item all correct  yields a reward of $r$.  The execution of $\j{Deliver}$ action with a wrong item, a wrong person, or a wrong room yield a penalty of $p_1$, $p_2$, $p_3$ each.
\[
\ba l
{\reward}\ r\ \iif\ \j{ItemReq}\mvis i\land \j{PersonReq}\mvis p \land \j{RoomReq}\mvis r\land\j{Deliver}(i, p, r)\land \sneg\j{Terminated},\\
{\reward}\ -\!\!p_1\ \iif\ \j{ItemReq}\mvis i\land \j{Deliver}(i', p', r')\land \sneg \j{Terminated}\ \ \ ( i \neq i'),\\
{\reward}\ -\!\!p_2\ \iif\ \j{PersonReq}\mvis p\land \j{Deliver}(i', p', r')\land \sneg \j{Terminated}\ \ \ ( p \neq p'),\\
{\reward}\ -\!\!p_3\ \iif\ \j{RoomReq}\mvis r\land \j{Deliver}(i', p', r')\land \sneg \j{Terminated}\ \ \ ( r \neq r').
\ea 
\]
\BOCC
\begin{align}
\nonumber  &{\reward}\ 10\ \iif\ \j{ItemReq}\mvis i\ \after\ \j{Deliver}(i, p, r)\land \sneg \j{Terminated},\\
\nonumber \ &{\reward}\ 10\ \iif\ \j{PersonReq}\mvis p\ \after\ \j{Deliver}(i, p, r)\land \sneg \j{Terminated},\\
\nonumber (4)\ &{\bf reward}\ 10\ \iif\ \j{RoomReq}\mvis r\ \after\ \j{Deliver}(i, p, r)\land \sneg \j{Terminated}.
\end{align}
\EOCC
Asking ``which item'' question when the actual item being requested is $i$ returns an item $i'$ as observation in accordance with the probability distribution defined by pf constant $\j{Pf\_WhichItem}(i)$, shown below. ``Which person'' and ``Which room'' questions are represented in a similar way. 
\beq 
\ba l
{\observed}\ \j{ItemObs}\mvis i'\ \iif\ \j{ItemReq}\mvis i\land\sneg\j{Terminated}\  \after\ \j{WhichItem}\land \j{Pf\_WhichItem}(i)\mvis i',\\
\caused\ \j{Pf\_WhichItem}(\j{Coffee}) \mvis  \{\j{Coffee}: 0.7, \j{Coke}: 0.1, \j{Cookies}: 0.1, \j{Burger}: 0.1\},\\
\caused\ \j{Pf\_WhichItem}(\j{Coke}) \mvis  \{\j{Coffee}: 0.1, \j{Coke}: 0.7, \j{Cookies}: 0.1, \j{Burger}: 0.1\},\\
\caused\ \j{Pf\_WhichItem}(\j{Cookies}) \mvis  \{\j{Coffee}: 0.1, \j{Coke}: 0.1, \j{Cookies}: 0.7, \j{Burger}: 0.1\},\\
\caused\ \j{Pf\_WhichItem}(\j{Burger}) \mvis  \{\j{Coffee}: 0.1, \j{Coke}: 0.1, \j{Cookies}: 0.1, \j{Burger}: 0.7\},
\ea 
\eeq{pf-whichitem}

\BOC

\begin{align}
 &{\observed}\ \j{ItemObs}\mvis i'\ \iif\ \j{ItemReq}\mvis i\land\sneg\j{Terminated}\  \after\ \j{WhichItem}\land \j{Pf\_WhichItem}(i)\mvis i',\\
\label{pf-whichitem-1} & \caused\ \j{Pf\_WhichItem}(\j{Coffee}) \mvis  \{\j{Coffee}: 0.7, \j{Coke}: 0.1, \j{Cookies}: 0.1, \j{Burger}: 0.1\},\\
\label{pf-whichitem-2} \ & \caused\ \j{Pf\_WhichItem}(\j{Coke}) \mvis  \{\j{Coffee}: 0.1, \j{Coke}: 0.7, \j{Cookies}: 0.1, \j{Burger}: 0.1\},\\
\label{pf-whichitem-3} & \caused\ \j{Pf\_WhichItem}(\j{Cookies}) \mvis  \{\j{Coffee}: 0.1, \j{Coke}: 0.1, \j{Cookies}: 0.7, \j{Burger}: 0.1\},\\
\label{pf-whichitem-4} & \caused\ \j{Pf\_WhichItem}(\j{Burger}) \mvis  \{\j{Coffee}: 0.1, \j{Coke}: 0.1, \j{Cookies}: 0.1, \j{Burger}: 0.7\},
\end{align}
\EOC

\BOCC
\NB{The following part can be removed}

\begin{align}
\nonumber &{\observed}\ \j{PersonObs}\mvis p'\ \iif\ \j{PersonReq}\mvis p\land\sneg\j{Terminated}\ \after\ \j{WhichPerson}\land \j{Pf\_WhichPerson(p)}\mvis p',\\
\nonumber & \caused\ \j{Pf\_WhichPerson}(\j{Bob}) \mvis  \{\j{Alice}: 0.8, \j{Bob}: 0.1, \j{Carol}: 0.1\},\\
\nonumber & \caused\ \j{Pf\_WhichPerson}(\j{Bob}) \mvis  \{\j{Alice}: 0.1, \j{Bob}: 0.8, \j{Carol}: 0.1\},\\
\nonumber & \caused\ \j{Pf\_WhichPerson}(\j{Carol}) \mvis  \{\j{Alice}: 0.1, \j{Bob}: 0.1, \j{Carol}: 0.8\},\\
\nonumber &{\observed}\ \j{ObsRoom}\mvis r'\ \iif\ \j{PersonReq}\mvis r\land\sneg\j{Terminated}\\
\nonumber &\ \ \ \ \ \ \after\ \j{AskPerson}\land \j{Pf\_WhichPerson(r)}\mvis r',\\
\nonumber & \caused\ \j{Pf\_WhichRoom}(R_1) \mvis  \{R_1: 0.8, R_2: 0.1, R_3: 0.1\},\\
\nonumber & \caused\ \j{Pf\_WhichRoom}(R_2) \mvis  \{R_1: 0.1, R_2: 0.8, R_3: 0.1\},\\
\nonumber & \caused\ \j{Pf\_WhichRoom}(R_3) \mvis  \{R_1: 0.1, R_2: 0.1, R_3: 0.8\}.
\end{align}
\NB{Explain why need terminated} 
\EOCC
When the robot asks the confirmation question ``is the item $i$?'', the human's answer could be sometimes mistakenly recognized, and the probability distribution of the answer depends on whether the item $i$ is indeed what the human asked for. We use two pf constants, $\j{Pf\_ConfirmWhenCorrect}$ and $\j{Pf\_ConfirmWhenIncorrect}$ to specify each of the probability distributions depending on whether the robot's guess is correct or not.  When the robot asks to confirm if the item requested is $i$, which is indeed what the human requested:
\begin{align}
\nonumber &{\observed}\ \j{Confirmation}\mvis v\ \iif\ \j{ItemReq}\mvis i\land\sneg\j{Terminated}\\
\nonumber &\hspace{3cm} \after\ \j{ConfirmItem}(i)\land \j{Pf\_ConfirmWhenCorrect} \mvis  v.  \qquad(v \in \{{\tt Yes}, {\tt No}\}) \\
\nonumber & \caused\ \j{Pf\_ConfirmWhenCorrect} \mvis  \{{\tt Yes}: 0.8, {\tt No}: 0.2\}.
\end{align}
When the robot asks to confirm if the requested item is $i'$ whereas the actual item the human requested is {\cred $i$}:
\begin{align}
\nonumber &\observed\ \j{Confirmation}\mvis v\ \iif\ \j{ItemReq}\mvis i\land\sneg\j{Terminated}\\
\nonumber  &\ \hspace{3cm} \after\ \j{ConfirmItem}(i')\land \j{Pf\_ConfirmWhenIncorrect} \mvis  v \qquad (i\neq i'),\\
\nonumber & \caused\ \j{Pf\_ConfirmWhenIncorrect} \mvis  \{{\tt Yes}: 0.2, {\tt No}: 0.8\}.
\end{align}
(The probability distributions of these pf constants do not have to be complementary.)
 
The formulations of person- and room-related questions are described similarly, and omitted from the paper.

\BOCC
\begin{align*}
 &{\observed}\ \j{Confirmation}\mvis v\ \iif\ \j{PersonReq}\mvis p\land\sneg\j{Terminated}\\
     &\hspace{2cm}  \after\ \j{ConfirmPerson}(p)\land     \
                                      \j{Pf\_ConfirmWhenCorrect} \mvis  v,\\
 &{\observed}\ \j{Confirmation}\mvis v\ \iif\ \j{PersonReq}\mvis p\land\sneg\j{Terminated}\\
 &\ \ \ \ \ \ \ \after\ \j{ConfirmPerson}(p')\land  \j{Pf\_ConfirmWhenIncorrect} \mvis  v\qquad (p\neq p'),\\
 &{\observed}\ \j{Confirmation}\mvis v\ \iif\ \j{RoomReq}\mvis r\land\sneg\j{Terminated}\\
 &\ \ \ \ \ \ \  \after\ \j{ConfirmRoom}(r)\land \j{Pf\_ConfirmWhenIncorrect} \mvis  v,\\
  &{\observed}\ \j{Confirmation}\mvis v\ \iif\ \j{RoomReq}\mvis r\land\sneg\j{Terminated}\ \\
 &\ \ \ \ \ \ \ \after\ \j{ConfirmRoom}(r')\land  \j{Pf\_ConfirmWhenIncorrect} \mvis  v \qquad  (r\neq r').
\end{align*}
\EOCC

Asking which-questions has a cost of $c_1$; asking confirmation-questions has a cost of $c_2$.
\begin{align}
\nonumber &{\reward}\ -\!\!c_1\ \iif\ \top\ \after\ \j{WhichItem}, & & {\reward}\ -\!\!c_2\ \iif\ \top\ \after\ \j{ConfirmItem}(i),\\
\nonumber &{\reward}\ -\!\!c_1\ \iif\ \top\ \after\ \j{WhichPerson},  & &{\reward}\ -\!\!c_2\ \iif\ \top\ \after\ \j{ConfirmPerson}(p),\\
\nonumber &{\reward}\ -\!\!c_1\ \iif\ \top\ \after\ \j{WhichRoom}, & & {\reward}\ -\!\!c_2\ \iif\ \top\ \after\ \j{ConfirmRoom}(r).
\end{align}

Finally, all regular fluents in this domain are inertial:
\[
   \inertial\ \j{rf}     \qquad (\j{rf}\in\{\j{ItemReq}, \j{PersonReq}, \j{RoomReq}, \j{Terminated}\}).
\]

In the following subsections, we illustrate the elaboration tolerance of the above $\pbcp$ action description.
It should be noted that using a vanilla POMDP method, manipulating states, actions, or observation functions requires  significant engineering efforts, and a developer frequently has to tune prohibitively a large number of parameters. i{\sc corpp} and this research aim to avoid that through probabilistic reasoning about actions. In this work, we move forward from i{\sc corpp} to shield a developer from the syntax or algorithms of POMDPs. 

\subsection{Elaboration 1: Unavailable items} \label{ssec:elab1}
When an item becomes unavailable for delivery, we can simply remove that item from the domains of relevant constants. For example, when $\j{Coke}$ becomes unavailable, we simply replace the pf constant declarations in~\eqref{pf-whichitem} with
\begin{align}
\nonumber & \caused\ \j{Pf\_WhichItem}(\j{Coffee}) \mvis  \{\j{Coffee}: 0.78, \j{Cookies}: 0.11, \j{Burger}: 0.11\},\\
\nonumber & \caused\ \j{Pf\_WhichItem}(\j{Cookies}) \mvis  \{\j{Coffee}: 0.11, \j{Cookies}: 0.78, \j{Burger}: 0.11\},\\
\nonumber & \caused\ \j{Pf\_WhichItem}(\j{Burger}) \mvis  \{\j{Coffee}: 0.11, \j{Cookies}: 0.11, \j{Burger}: 0.78\}.
\end{align}

\subsection{Elaboration 2: Reflecting personal preference in reward function} \label{ssec:elab2}
We use a rigid fluent $\j{Interchangeable}(p, i_1, i_2)$ with the integer domain to represent to what degree the two items $i_1, i_2$ are interchangeable for person $p$. For  example, Alice does not mind when the robot delivers coke while she actually ordered coffee but she does mind when the robot delivers burger instead of coffee.
We add the following elaboration to represent object interchangeability.

\[
\ba l 
\caused\ \j{Interchangeable}(\j{Alice}, \j{Coffee}, \j{Coke}) \mvis  5,\\
\caused\ \j{Interchangeable}(\j{Alice}, \j{Coffee}, \j{Cookies}) \mvis  1,\\
\caused\ \j{Interchangeable}(\j{Alice}, \j{Coffee}, \j{Burger}) \mvis  -3.
\ea 
\]


%
We add the following causal law to reflect the interchangeability of the items.
\begin{align}
\nonumber &{\reward}\ x\ \iif\ \j{ItemReq}\mvis i\land\j{Interchangeable}(p, i, i')\mvis x\land \j{PersonReq}(p) \ \after\ \j{Deliver}(i', p', r').
\end{align}

Such knowledge can be used to enable the robot to be more conservative in delivering items, such as \emph{burger}, due to their low interchangeability with other items.

\subsection{Elaboration 3: Changing Perception Model} \label{ssec:elab3}
The speech recognition system may have different accuracies depending on the environment. For example, when there is background noise, its accuracy could drop. In this case, we can update the probability distribution for the relevant pf constant, controlled by auxiliary constants indicating the situation. We introduce a rigid constant called $\j{Noise}$, and then replace \eqref{pf-whichitem} with
\begin{align}
\nonumber & \caused\ \j{Pf\_WhichItem}(\j{Coffee}) \mvis  \{\j{Coffee}: 0.7, \j{Coke}: 0.1, \j{Cookies}: 0.1, \j{Burger}: 0.1\}\ {\bf unless}\ ab\\
\nonumber & \caused\ \j{Pf\_WhichItem}(\j{Coke}) \mvis  \{\j{Coffee}: 0.1, \j{Coke}: 0.7, \j{Cookies}: 0.1, \j{Burger}: 0.1\}\ {\bf unless}\ ab\\
\nonumber & \caused\ \j{Pf\_WhichItem}(\j{Cookies}) \mvis  \{\j{Coffee}: 0.1, \j{Coke}: 0.1, \j{Cookies}: 0.7, \j{Burger}: 0.1\}\ {\bf unless}\ ab\\
\nonumber & \caused\ \j{Pf\_WhichItem}(\j{Burger}) \mvis  \{\j{Coffee}: 0.1, \j{Coke}: 0.1, \j{Cookies}: 0.1, \j{Burger}: 0.7\}\ {\bf unless}\ ab \\
\end{align}
to make them defeasible.
We then define the probability distribution to override the original ones when there is loud background noise.
\begin{align}
\nonumber  & \caused\ \j{Pf\_WhichItem}(\j{Coffee}) \mvis  \{\j{Coffee}: \frac{6}{10}, \j{Coke}: \frac{4}{30}, \j{Cookies}: \frac{4}{30}, \j{Burger}: \frac{4}{30}\}\ \iif\ \j{Noise},\\
\nonumber  & \caused\ \j{Pf\_WhichItem}(\j{Coke}) \mvis  \{\j{Coffee}: \frac{4}{30}, \j{Coke}: \frac{6}{10}, \j{Cookies}: \frac{4}{30}, \j{Burger}: \frac{4}{30}\}\ \iif\ \j{Noise},\\
\nonumber  & \caused\ \j{Pf\_WhichItem}(\j{Cookies}) \mvis  \{\j{Coffee}: \frac{4}{30}, \j{Coke}: \frac{4}{30}, \j{Cookies}: \frac{6}{10}, \j{Burger}: \frac{4}{30}\}\ \iif\ \j{Noise},\\
\nonumber  & \caused\ \j{Pf\_WhichItem}(\j{Burger}) \mvis  \{\j{Coffee}: \frac{4}{30}, \j{Coke}: \frac{4}{30}, \j{Cookies}: \frac{4}{30}, \j{Burger}: \frac{6}{10}\}\ \iif\ \j{Noise}.
\end{align}
We add
\[
\caused\ ab\ \iif\ \j{Noise}
\]
to indicate that by default there is no background noise. When the robot agent detects that there is background noise, we add
\[
\caused\ \j{Noise}
\]
to the action description to update the generated POMDP to incorporate the new speech recognition probabilities. It should be noted that the speech recognition component is generally unreliable, though background noise further reduces its reliability. 

\BOCC
{\bf Elaboration 4: Incorporate commonsense in State Estimation}
Some knowledge about the environment can help the robot agent make more accurate decision when the speech recognition is not accurate. For example, at morning time people may be more likely to order coffee. To incorporate this, we introduce a rigid fluent $\j{Morning}$ with Boolean domain, indicating if it is morning. Then we can simply replace (26) with
\begin{align}
\nonumber &\caused\ \j{Init\_ItemReq} \mvis  \{\j{Coffee}: 0.25, \j{Coke}: 0.25, \j{Cookies}: 0.25, \j{Burger}: 0.25\}\ {\bf unless}\ \j{Morning}
\end{align}
to make it defeasible.
Then we add
\begin{align}
\nonumber &\caused\ \j{Init\_ItemReq} \mvis  \{\j{Coffee}: 0.4, \j{Coke}: 0.2, \j{Cookies}: 0.2, \j{Burger}: 0.2\}\ {\iif}\ \j{Morning}
\end{align}
\EOCC

\section{System {\sc pbcplus2pomdp}} \label{sec:system}

We implemented the prototype system {\sc pbcplus2pomdp}, which takes a $\pbcp$ action description $D$ as input and outputs the POMDP $M(D)$ in the input language of the POMDP solver {\sc APPL}.\footnote{\url{http://bigbird.comp.nus.edu.sg/pmwiki/farm/appl/}} The system uses {\sc lpmln2asp} \cite{lee17computing} with exact inference on $D_1$ and $D_0$ to generate the components of POMDP: all states, all actions, all transitions and their probabilities, all observations and their probabilities and transition rewards as defined in Definition~\ref{def:pbcp-pomdp}. The system is publicly available at \url{https://github.com/ywang485/pbcplus2pomdp}, along with several examples.

Even though we limit the computation to $D_0$ and $D_1$, i.e., at most one step action execution is considered, the number of stable models may become too large to enumerate all. Since the transition probabilities, rewards, observation probabilities are per each action, the system implements a compositional way to generate the POMDP model by partitioning the actions in different groups and generating the POMDP model per each group by omitting the causal laws involving other actions and their pf constants. This ``compositional" mode often saves the POMDP generation time drastically.\footnote{The more detailed description of the algorithm is given in~\ref{sec:compo} of the supplementary material corresponding 
to this paper at the TPLP archives.
}


\BOCC
{\cred 
In particular, the inputs of {\sc pbcplus2pomdp(comp)} include the following: 
\begin{itemize}
    \item $\lpmln$ program $\Pi(m)$, parameterized with maximum timestep $m$, that contains $\lpmln$ translation of fluent dynamic laws, observation dynamic laws and utility laws with no occurrence of action constant, and static laws, as well as pf constant declarations of pf constants that occur in those causal laws (see Fig. ~\ref{fig:pbcplus-causal-laws});
    \item For each group of actions $a_i\in$ in $a_1$,$\dots$, $a_n$, an $\lpmln$ program $\Pi_i(m)\cup C_i(m)$, parameterized with maximum timestep $m$; $\Pi_i(m)$ contains translation of fluent dynamic laws, observation dynamic laws and utility laws where only actions in $a_i$ can occur in the body, as well as pf constant declarations of pf constants that occurs in those causal laws; $C_i(m)$ contains choice rules (possibly with cardinality bounds) to generate exactly one action in the group $a_i$; It is up to the user how to group the actions; 
    \item Discount factor.
\end{itemize}
The system outputs the POMDP definition $M(D)$, so that $D_m=\Pi(m)\cup \Pi_1(m)\cup\dots\Pi_n(m)\cup C(m)$, where $C(m)$ is the choice rule with cardinality constraint to generate at most one action in $a_1, \dots, a_n$ for each timestep $i\in\{0, \dots, m-1\}$. The transition probabilities, observation probabilities and reward function of $M(D)$ are obtained by conjoining those from each of $\Pi\cup \Pi_i\cup C_i$ ($i\in\{1, \dots, n\}$). 

Formally, let ${\bf S}$, $\Omega$, $P_{M(D)}$, $O_{M(D)}$, $R_{M(D)}$ be the set of states, the set of observations, transition probabilities, observation probabilities and reward function of $M(D)$, resp. system {\sc pbcplus2pomdp} calls {\sc lpmln2asp} first to solve $\Pi(0)$ to obtain ${\bf S}$, and then $\Pi(1)\cup\Pi_i(1)\cup C_i(1)$ to obtain $P_{M(D)}$, $O_{M(D)}$, $R_{M(D)}$ as follows:
\[
P_{M(D)}(s, a, s') = P_{\Pi(1)\cup \Pi_i(1)\cup C_i(1)}(1:s' \mid 0:s, 0:a)
\]
\[
O_{M(D)}(s, a, o) = P_{\Pi(1)\cup \Pi_i(1)\cup C_i(1)}(1:o\mid 1:s, 0:a)
\]
\[
R_{M(D)}(s, a, s') = E[U_{\Pi(1)\cup \Pi_i(1)\cup C_i(1)}(0:s, 0:a, 1:s')]
\]
for each $a\in a_i$, $s, s'\in {\bf S}$ and $o\in\Omega$.
}

\EOCC 

\BOC
In particular, the inputs of {\sc pbcplus2pomdp} include the following: 
\begin{itemize}
    \item $\lpmln$ program $\Pi$ that contains no fluent dynamic laws or dynamic observation laws except the ones saying fluents follow the common sense law of inertia;
    \item For each group of actions $a_i$ in $a_1$,$\dots$, $a_n$, an $\lpmln$ program $\Pi_i\cup C_i$; $\Pi_i$ contains translation of dynamic observation causal laws and dynamic observation laws with actions only in $a_i$ occurs in the body, as well as definitions of pf constants that occurs in those causal laws; $C_i$ contains choice rules to generate exactly one action in the group $a_i$; It is up to the user how to group the actions; 
    \item Discount factor.
\end{itemize}
\EOC

\section{Evaluation}  \label{sec:evaluation}

All experiments reported in this section were performed on a machine powered by 4 Intel(R) Core(TM) i5-2400 CPU with OS Ubuntu 14.04.5 LTS and 8G memory.

\BOC
\begin{table}[t]
\caption{Running Statistics of {\sc pbcplus2pomdp} System}
\label{fig:system-analysis}
\includegraphics[width=0.99\textwidth]{pbcplus2pomdp-statistics.png}
\end{table}
\EOC

\subsection{Evaluation of Planning Efficiency}


\begin{table}\addtolength{\itemsep}{-5mm}
{\footnotesize
\centering
\begin{tabular}{ c  c  c  c  c  c } 
\hline
 & \multicolumn{2}{c}{POMDP Generation Time} & \multicolumn{3}{c}{POMDP Solving Time ({\sc APPL})} \\ \hline
 Domain Size & {\sc pbcplus2pomdp} & {\sc pbcplus2pomdp} & $\gamma=0.9$ & $\gamma = 0.8$ & $\gamma = 0.7$  \\ 
   & (naive) & (compo)  &  & \\  \hline
$\begin{array}{c}2i2p2r\\ \#states=16\\ \#actions=18\\ \#observations=9\end{array}$ & 49m10.495s & 0m13.611s & 0m6.123s & 0m0.680s & 0m0.249s \\ \hline
$\begin{array}{c}2i3p2r\\\#states=24\\ \#actions=23\\ \#observations=10\end{array}$ & $>$ 1hr & 0m22.723s & 4m43.572s  & 0m21.939s & 0m2.294s \\ \hline
$\begin{array}{c}3i3p2r\\\#states=36\\ \#actions=30\\ \#observations=11\end{array}$ & $>$ 1hr & 0m41.944s & $>$ 1hr & 8m14.415s & 0m37.944s \\ \hline
$\begin{array}{c}4i3p2r\\\#states=48\\ \#actions=37\\ \#observations=12\end{array}$ & $>$ 1hr & 2m56.652s & $>$ 1hr & $>$ 1hr & 10m50.248s \\ \hline
\end{tabular}
}\\[-1em]
\caption{Running Statistics of POMDP Model Generation and Solving in Dialog Example}
\label{tab:statistics-efficiency}
\end{table}

We report the running statistics of POMDP generation with our {\sc pbcplus2pomdp} system and POMDP planning with {\sc APPL} on the dialog example (as described in Section \ref{sec:elab}) in 
Table~\ref{tab:statistics-efficiency}.  We test domains with different numbers of items, people, and rooms. {\sc pbcplus2pomdp(naive)} generates POMDP in a non-compositional way while {\sc pbcplus2pomdp(compo)} generates POMDP in a compositional way (as described in Section \ref{sec:system}) by partitioning actions into
   $\{\j{ConfirmItem}(i)\mid i\in \j{Item}\}$,
   $\{\j{ConfirmPerson}(p)\mid p\in \j{Person}\}$,
   $\{\j{ConfirmRoom}(r)\mid r\in \j{Room}\}$,
   $\{\j{WhichItem}\}$,
   $\{\j{WhichPerson}\}$,
   $\{\j{WhichRoom}\}$,
   $\{\j{Deliver}(i, p, r)\mid i\in \j{Item}, p\in \j{Person}, r\in \j{Room}\}$.

$\gamma$ is a discount factor. ``POMDP solving time (APPL)'' refers to the running time of {\sc APPL} until the convergence to a target precision of $0.1$. The {\sc pbcplus2pomdp(compo)} mode is much more efficient than the {\sc pbcplus2pomdp(naive)} mode for the dialog domain. 

\BOC
We report the performance of the system on tiger example with the increased number of tigers\footnote{The number of doors is alway number of tigers plus 1.} in Fig.~\ref{fig:system-analysis}. The discount factor is set to be $0.9$.
\EOC

\renewcommand\baselinestretch{1}\selectfont

\subsection{Evaluation of Solution Quality}

$\pbcp$ provides a high-level description of POMDP models such that various elaborations on the underlying action domain can be easily achieved by changing a small part of the $\pbcp$ action description, whereas such elaboration would require a complete reconstruction of transition/reward/observation matrices at POMDP level. 
In Sections \ref{ssec:elab1}, \ref{ssec:elab2} and \ref{ssec:elab3}, we have illustrated this point with the three example elaborations. In this subsection, we evaluate the impact of the three elaborations on dynamic planning, in the sense that the low-level POMDP (planning module) can be updated automatically once the high-level $\pbcp$ action description (reasoning module) detects changes in the environment to generate better plans. For each of the thee elaborations, we compare the plan generated from a static POMDP that does not reflect environmental changes, and the one generated from the adaptive POMDP that is updated by $\pbcp$ reasoning to reflect environmental changes.

\begin{figure}[t]
\centering
		\includegraphics[width=0.95\textwidth]{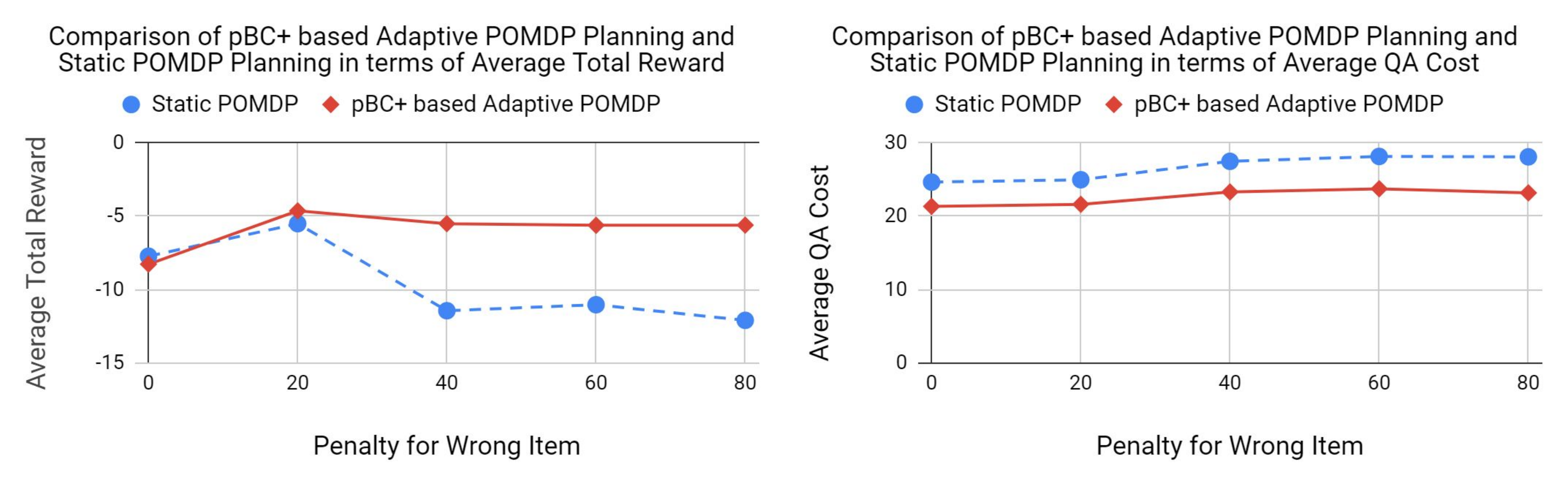}
	\caption{Impact of Elaboration 1 on Policy Generated}
	\label{fig:eb1-evalation}
\end{figure}

Fig.~\ref{fig:eb1-evalation} compares the policies generated from the static POMDPs (baseline) and from the POMDP dynamically generated using $\pbcp$, where the two items of burger and cookies might be unavailable (Elaboration 1). 
We have run $1000$ simulation trials. The diagram on the left compares them in terms of average total reward from the simulation runs, and the right is in terms of average QA cost (accumulated penalty by asking questions). 
In this experiment, the discount factor is $0.95$ (which offers the dialog agent a relatively long horizon), $c_1$ is $4.0$, $c_2$ is $2.0$, $r$ is $20.0$, $p_2$ is $20.0$, and $p_3$ is $30.0$. 
Action policies are generated using APPL in at most $120$ seconds. 
We observe that the adaptive POMDP (ours) achieves a higher average total reward when the penalty for the wrong item is positive, and the adaptive POMDPs are able to complete deliveries with less QA costs. It is worth noting that by reflecting unavailable items, $\pbcp$ reduces the size of the generated POMDP models, resulting in shorter POMDP-solving times. As can be seen from Table~\ref{tab:statistics-efficiency}, for a domain that contains $2$ items, $3$ people and $2$ rooms, POMDP generation plus POMDP solving takes way less time than POMDP solving on a domain with $4$ items, $3$ people and $2$ rooms.

\begin{figure}
\centering
		\includegraphics[width=0.75\textwidth]{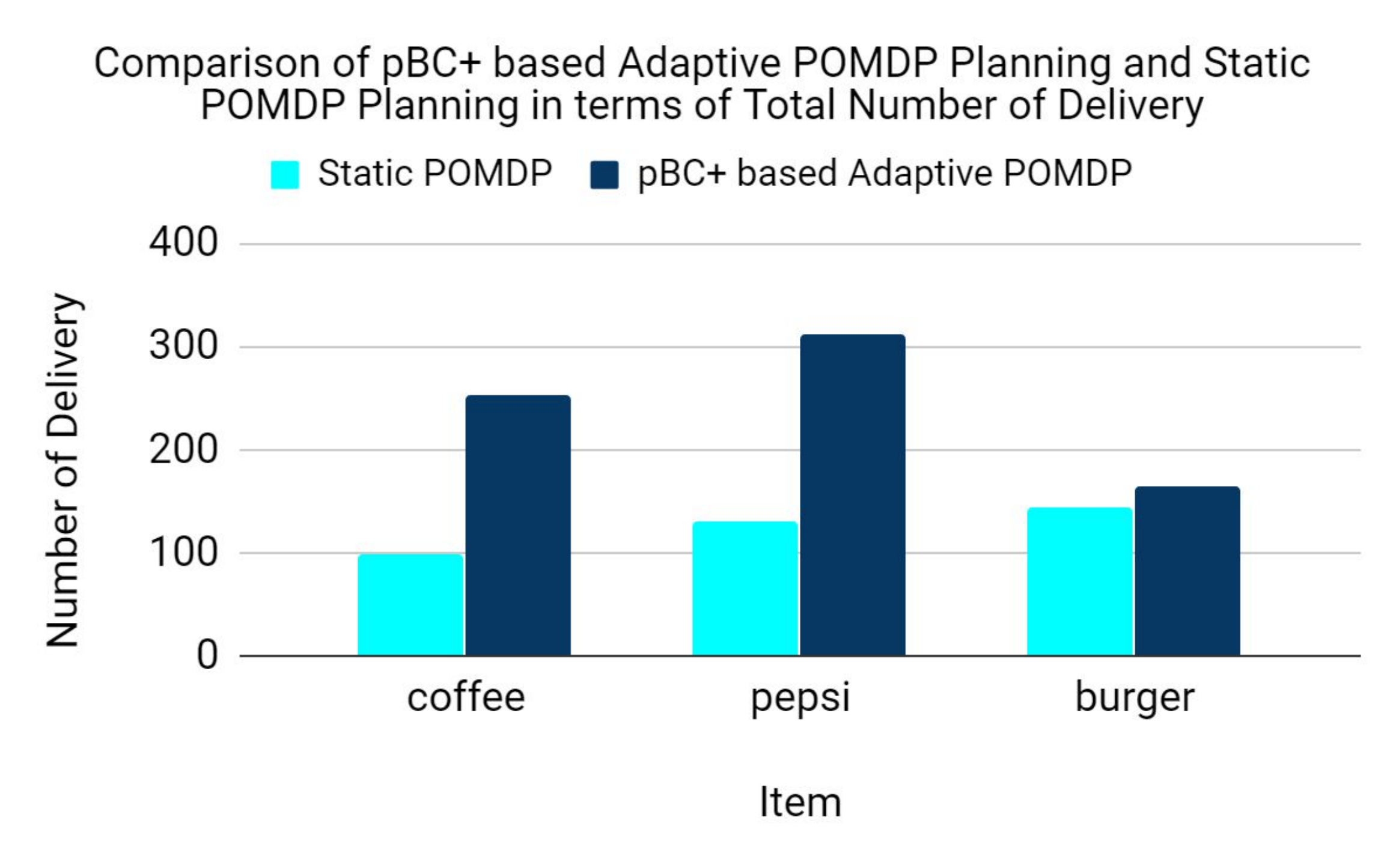} 
	\caption{Impact of Elaboration 2 on Policy Generated}
	\label{fig:eb2-evalation}
\end{figure}

Fig.~\ref{fig:eb2-evalation} compares the policies generated from the static POMDP and from $\pbcp$ based adaptive POMDP when item interchangeability is introduced (Elaboration 2). We replaced cookies with pepsi in the domain, added causal laws to indicate that when coke is being requested, delivering pepsi instead yields a reward of $15$, delivering coffee instead yields a reward of $5$ and delivering burger instead yields an additional penalty of $20$ (in the presence of penalty $p_1$). We have run $10000$ simulations, and for all of the simulations, the actual item being requested is fixed to be coke.\footnote{The item is fixed to be coke only during simulation, not during policy generation.} For the static POMDP, $9628$ deliveries were correct, and for the adaptive POMDP, $9270$ deliveries were correct. Note that although the static POMDP achieves more correct deliveries, the dynamically generated POMDPs (our approach) achieved higher average total reward by asking fewer questions. 
The policy generated from the static POMDP gives similar numbers of deliveries for each item that is not coke, while the policy generated from the adaptive POMDP delivered pepsi the most and burger the least, which is aligned with our setting of interchangeability. The discount factor for this experiment is set to be $0.99$. $c_1$ is $6$, $c_2$ is $4$, $r$ is $5$, $p_1$ is $5$, $p_2$ is $20$ and $p_3$ is $30$. Policies from both POMDPs are generated by APPL with $120$ seconds.

\begin{figure}[t]
\centering
		\includegraphics[width=\textwidth]{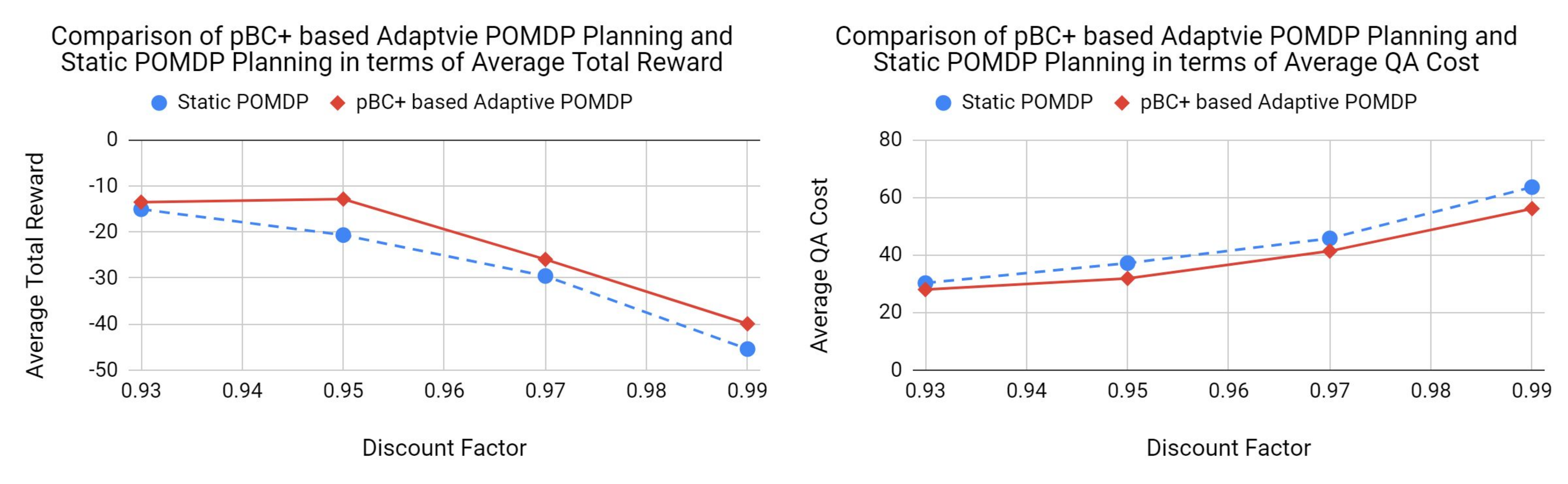}
	\caption{Impact of Elaboration 3 on Policy Generated}
	\label{fig:eb3-evalation}
\end{figure}

Fig.~\ref{fig:eb3-evalation} compares the policies generated from the static POMDP and from $\pbcp$ based adaptive POMDP when there is a background noise (Elaboration 3). To reflect environmental noise, we lowered the observation probability of correct answers by $0.1$ (and the remaining answers are uniformly distributed). We have run $1000$ simulations. The diagram on the left compares them in term of average total reward from the simulation runs, and the diagram on the right compares them in terms of average QA cost (accumulated cost from questions asked) from the simulation runs. In this experiment, $c_1$ is $4$, $c_2$ is $2$, $r$ is $20$, $p_2$ is $20$ and $p_3$ is $30$. Policies from both POMDPs are generated by APPL with $120$ seconds. It can be seen from the diagrams that while the average total reward of both POMDPs decreases as the discount factor increases, the adaptive POMDP achieves higher average total reward by asking fewer questions.

\BOC
{\cblu[[ how to measure elaboration tolerance in comparison with shiqi's earlier work]]
b
\begin{itemize}
    \item[E1:] Compare pBC+ with standard POMDP in the total policy generation time, where standard POMDP approach requires larger POMDP models to account for the possible domain changes. For pBC+, the policy generation time includes POMDP construction time as well. We can show that, given limited policy generation time, pBC+ generates better-quality policies. 
    \item[E2:] We just qualitatively compare pBC+ with i{\sc corpp}, saying something like i{\sc corpp} uses P-log that does not support numerical reasoning, whereas pBC+ is a unified representation that supports logical-probabilistic reasoning as well as numerical computations. 
    \item[E3:] Again, the i{\sc corpp} paper did not mention probabilistic reasoning about observations. So, we can qualitatively make some comparisons. If we want more quantitative results, we can compute the arithmetic mean over the two ``noise'' conditions, and refer this as the i{\sc corpp} approach. Then we can generate curves similar to those in Fig. 3 of the CORPP paper (AAAI'15). 
\end{itemize}}
\EOC


\section{Related Work} \label{sec:related}

Intelligent agents need the capabilities of both reasoning about declarative knowledge, and probabilistic planning toward achieving long-term goals. 
A variety of algorithms have been developed to integrate commonsense reasoning and probabilistic planning~\cite{hanheide2017robot,zhang15mixed,zhang15corpp,sridharan2019reba,chitnis2018integrating,zhang17dynamically,amiri2018multi,veiga2019hierarchical}, and some of them, such as~\cite{sridharan2019reba} and~\cite{amiri2018multi}, also include non-deterministic dynamic laws for observations. 
Although the algorithms use very different computational paradigms for representing and reasoning with human knowledge (e.g., logics, probabilities, graphs, etc), they all share the goal of leveraging declarative knowledge to improve the performance in probabilistic planning. In these works, the hypothesis is that human knowledge potentially can be useful in guiding robot behaviors in the real world, while the challenge is that human knowledge is sparse, incomplete, and sometimes unreliable. 
In this research, we share the same goal of utilizing contextual knowledge from people to help intelligent agents in sequential decision-making tasks while accounting for the uncertainty in perception and action outcomes. 

Among the algorithms that integrate commonsense reasoning and probabilistic planning paradigms, i{\sc corpp} enabled an agent to reason with contextual knowledge to dynamically construct complete probabilistic planning models~\cite{zhang17dynamically} for adaptive robot control, where P-log was used for logical-probabilistic reasoning~\cite{baral09probabilistic}. 
Depending on the observability of world states, i{\sc corpp} uses either Markov Decision Processes (MDPs)~\cite{puterman2014markov} or Partially Observable MDPs (POMDPs)~\cite{kaelbling1998planning} for probabilistic planning. 
As a result, i{\sc corpp} has been applied to robot navigation, dialog system, and manipulation tasks~\cite{zhang17dynamically,amiri2018multi}. 
In this work, we develop a unified representation and a corresponding implementation for i{\sc corpp}, where the entire reasoning and planning system can be encoded using a single program, and practitioners are completely shielded from the technical details of formulating and solving (PO)MDPs. 
In comparison, i{\sc corpp} requires significant engineering efforts (e.g., using Python or C++) for ``gluing'' the computational paradigms used by the commonsense reasoning and probabilistic planning components. 

Recently, researchers have developed algorithms to incorporate knowledge representation and reasoning into reinforcement learning (RL)~\cite{sutton2018reinforcement}, where the goal is to provide the learning agents with guidance in action selections through reasoning with declarative knowledge. 
Notable examples include~\cite{leonetti2016synthesis,yang2018peorl,abs-1811-08955,abs-1809-11074,lyu2019sdrl,kimAAAI2019}. 
In this research, we assume the availability of world models, including both states and dynamics, in a declarative form. 
In case of world models being unavailable, incomplete, or dynamically changing, there is the potential of combining the above ``knowledge-driven RL'' algorithms, particularly the ones using model-based RL such as~\cite{abs-1809-11074}, with our new representation to enable agents to simultaneously learn and reason about world models to compute action policies. 

\BOC
{\cred 
Knowledge representation and reasoning paradigms have been developed to support reasoning with both logical and probabilistic knowledge [LPMLN, MLN, PSL, P-log...]. 
These declarative languages and systems have been applied to a variety of reasoning tasks. 
Intelligent agents (such as robots) frequently face tasks that require more than one action so as to achieve long-term goals, where reasoning about actions becomes necessary. 

Complex sequential decision-making has been decomposed into the two sub-tasks of commonsense reasoning and probabilistic planning~\cite{zhang15corpp,zhang15mixed}
. 
In that line of research, a reasoner was used for state estimation with contextual knowledge and a planner focuses on selecting actions to maximize long-term reward. 
More recently, logical-probabilistic knowledge has been used for reasoning about both the current state and the dynamics of the world, resulting in an algorithm called i{\sc corpp}~\cite{zhang17dynamically}. 
i{\sc corpp} builds on two computational paradigms of P-log~\cite{baral09probabilistic}
and (PO)MDPs~\cite{kaelbling1998planning}
for reasoning and planning respectively, and bridging the gap requires significant engineering work. 
In comparison, the semantics of pBC+ and its reasoning system together support the direct generation of planning models, which can be further used for computing action policies using off-the-shelf (PO)MDP solvers. 
}
\EOC

{\cblu 
In an earlier work \cite{tran04encoding}, the authors show how Pearl's probabilistic causal model can be encoded in a probabilistic action language PAL \cite{baral02reasoning}. 
}

\section{Conclusion and Future Work} \label{sec:conclusion}

In this paper, we present a principled way of integrating probabilistic logical reasoning and probabilistic planning. This is done by extending  probabilistic action language $\pbcp$ \cite{lee18aprobabilistic,wang19elaboration} to be able to express utility, belief states, and observation as in POMDP models. Inheriting the advantages of action languages, the new action language provides an elaboration tolerant representation of POMDP that is convenient to encode commonsense knowledge. 

{\cblu 
One of the well known problems limiting applications of POMDPs is sensitivity of the optimal behavior to the small changes in the reward function and the probability distribution. Because of this sensitivity care must be taken in choosing the reward function as well as the probability distribution. The choice of these, and especially of the latter is a non-trivial problem, which is outside of the scope of the paper. 
POMDP algorithms perform poorly in scalability in many applications. Although the language and system developed in this paper can potentially alleviate this issue, we believe this is a challenging problem that deserves more effort, and we leave it to future work. 
}

The current prototype implementation is not highly scalable when the number of transitions becomes large. For a more scalable generation of the POMDP input using the $\lpmln$ system, we could use the sampling method in $\lpmln$ inference, which we leave for future work.




\bigskip
\noindent
{\bf Acknowledgements:} We are grateful to the anonymous referees for their useful comments. The first and the third author's work was partially supported by the National Science Foundation under Grant IIS-1815337.

\bibliographystyle{acmtrans}

\newpage

\setcounter{page}{1}
\title{Appendix: Bridging Commonsense Reasoning and Probabilistic Planning via a Probabilistic Action Language}

\begin{center}

{\large\textnormal{Online appendix for the paper}}   \\
\medskip
{\Large {\sl Bridging Commonsense Reasoning and Probabilistic Planning via a Probabilistic Action Language}
\\
\medskip
{\large\textnormal{published in Theory and Practice of Logic Programming}}
}

\medskip
Yi Wang$^*$, Shiqi Zhang$^\#$, Joohyung Lee$^*$ \\
$^*$Arizona State University, USA \ \ \  $^\#$ SUNY Binghamton, USA

\end{center}

\thispagestyle{empty}

\begin{appendix}

\section{Extended Review of Preliminaries}

\subsection{Review: Language $\lpmln$}

 An $\lpmln$ program is a finite set of weighted rules $w: R$ where $R$ is a rule and $w$ is a real number (in which case, the weighted rule is called {\em soft}) or $\alpha$ for denoting the infinite weight (in which case, the weighted rule is called {\em hard}). Throughout the paper, we assume that the language is propositional. Schematic variables can be introduced via grounding as usual in answer set programming. 
 
For any $\lpmln$ program $\Pi$ and any interpretation~$I$, 
$\overline{\Pi}$ denotes the usual (unweighted) ASP program obtained from $\Pi$ by dropping the weights, and
${\Pi}_I$ denotes the set of $w: R$ in $\Pi$ such that $I\models R$.

In general, an $\lpmln$ program may even have stable models that violate some hard rules, which encode definite knowledge. However, throughout the paper, we restrict attention to $\lpmln$ programs whose stable models do not violate hard rules. 
More precisely, given a ground $\lpmln$ program $\Pi$, $\sm[\Pi]$ denotes the set
\[
\ba l
\{I\mid \text{$I$ is a (deterministic) stable model of $\Pi_I$ that satisfies all hard rules in $\Pi$} \} .
\ea
\]
The weight of an interpretation $I$, denoted $W_{\Pi}(I)$, is defined as 
\[
 W_\Pi(I) =
\begin{cases}
  exp\Bigg(\sum\limits_{w:R\;\in\; {\Pi}_I} w\Bigg) & 
      \text{if $I\in\sm[\Pi]$}; \\
  0 & \text{otherwise},
\end{cases}
\]
and the probability of $I$, denoted $P_\Pi(I)$, is defined as
\[
  P_\Pi(I)  = 
  \frac{W_\Pi(I)}{\sum\limits_{J\in {\rm SM}[\Pi]}{W_\Pi(J)}}. 
\]


\subsection{Review: $\dtlpmln$}

We extend the syntax and the semantics of $\lpmln$ to $\dtlpmln$ by introducing atoms of the form 
\begin{equation}\label{eq:utility-atoms}
{\tt utility}(u, {\bf t})
\end{equation}
where $u$ is a real number, and ${\bf t}$ is an arbitrary list of terms. These atoms can only occur in the head of hard rules of the form
\begin{equation}\label{eq:utility-rule}
\alpha: {\tt utility}(u, {\bf t}) \leftarrow \j{Body}
\end{equation}
where $\j{Body}$ is a list of literals. 
We call these rules {\em utility rules}.

The weight and the probability of an interpretation are defined the same as in $\lpmln$.
The {\em utility} of an interpretation $I$ under $\Pi$ is defined as
\[
U_{\Pi}(I) = \underset{{\tt utility}(u, {\bf t})\in I}{\sum} u .
\]
The {\em expected utility} of a proposition $A$ is defined as
\begin{equation}\label{eq:expected-utility}
E[U_{\Pi}(A)] = \underset{I\models A}{\sum}\  
    U_{\Pi}(I) \times P_{\Pi}(I\mid A) . 
\end{equation}

\subsection{Review: Multi-Valued Probabilistic Programs} \label{ssec:mvpp}
Multi-valued probabilistic programs \cite{lee16weighted} are a simple fragment of $\lpmln$ that allows us to represent probability more naturally. 

We assume that the propositional signature $\sigma$ is constructed from ``constants'' and their ``values.'' 
A {\em constant} $c$ is a symbol that is associated with a finite set $\j{Dom}(c)$, called the {\em domain}. 
The signature $\sigma$ is constructed from a finite set of constants, consisting of atoms $c\!=\!v$~\footnote{%
Note that here ``='' is just a part of the symbol for propositional atoms, and is not  equality in first-order logic. }
for every constant $c$ and every element $v$ in $\j{Dom}(c)$.
If the domain of~$c$ is $\{\false,\true\}$ then we say that~$c$ is {\em Boolean}, and abbreviate $c\mvis\true$ as $c$ and $c\mvis\false$ as~$\sneg c$. 

We assume that constants are divided into {\em probabilistic} constants and {\em non-probabilistic} constants.
A multi-valued probabilistic program ${\bf \Pi}$ is a tuple $\langle \j{PF}, \Pi \rangle$, where
\begin{itemize}
\item $\j{PF}$ contains \emph{probabilistic constant declarations} of the following form:
\begin{equation}\label{eq:probabilistic-constant-declaration}
p_1::\ c\mvis v_1\mid\dots\mid p_n::\ c\mvis v_n
\end{equation}
one for each probabilistic constant $c$, where $\{v_1,\dots, v_n\}=\j{Dom}(c)$, $v_i\ne v_j$, $0\leq p_1,\dots,p_n\leq1$ and $\sum_{i=1}^{n}p_i=1$. We use $M_{\bf \Pi}(c=v_i)$ to denote $p_i$.
In other words, $\j{PF}$ describes the probability distribution over each ``random variable''~$c$. 

\item $\Pi$ is a set of rules such that the head contains no probabilistic constants.
\end{itemize}

The semantics of such a program ${\bf \Pi}$ is defined as a shorthand for $\lpmln$ program $T({\bf \Pi})$ of the same signature as follows.
\begin{itemize}
\item For each probabilistic constant declaration (\ref{eq:probabilistic-constant-declaration}), $T({\bf \Pi})$ contains, 
for each $i=1,\dots, n$,
(i) $ln(p_i):  c\mvis v_i$  if $0<p_i<1$; 
(ii) $\alpha:\ c\mvis v_i$ if $p_i=1$;
(iii) $\alpha:\ \bot\ar c\mvis v_i$ if $p_i=0$.

\item  For each rule $\j{Head}\ar\j{Body}$ in $\Pi$, $T({\bf \Pi})$ contains
$
\alpha:\ \  \j{Head}\ar\j{Body}. 
$

\ii For each constant $c$, $T({\bf \Pi})$ contains the uniqueness of value constraints
\beq
\ba {rl}
   \alpha: & \bot \ar c\mvis v_1\land c=v_2 
\ea 
\eeq{uc}
for all $v_1,v_2 \in\j{Dom}(c)$ such that $v_1\ne v_2$, and the existence of value constraint
\beq
\ba {rl}
  \alpha: & \bot \ar \neg \bigvee\limits_{v \in {Dom}(c)} c\mvis v\ .
\ea 
\eeq{ec}
\end{itemize}

In the presence of the constraints \eqref{uc} and \eqref{ec}, assuming $T({\bf \Pi})$ has at least one (probabilistic) stable model that satisfies all the hard rules, a (probabilistic) stable model $I$ satisfies $c=v$ for exactly one value $v$, so we may identify $I$ with the value assignment that assigns $v$ to $c$.

\subsection{Review: Action Language $\pbcp$ with Utility}

\subsubsection{Syntax of $p\cal{BC}+$}
We assume a propositional signature~$\sigma$ as defined in Section~\ref{ssec:mvpp}.
We further assume that the signature of an action description is divided into four groups: {\em fluent constants}, {\em action constants},  {\em pf (probability fact) constants} and {\em  initpf (initial probability fact) constants}. Fluent constants are further divided into {\em regular} and {\em statically determined}. The domain of every action constant is Boolean. 
A {\em fluent formula} is a formula such that all constants occurring in it are fluent constants. 

The following definition of $p\cal{BC}$+ is based on the definition of ${\cal BC}$+ language from \cite{babb15action1}.

A {\em static law} is an expression of the form
\begin{equation}\label{eq:static-law}
\caused\ F\ \iif\ G
\end{equation}
where $F$ and $G$ are fluent formulas.


A {\em fluent dynamic law} is an expression of the form
\begin{equation}\label{eq:fluent-dynamic-law}
\caused\ F\ \iif\ G\ \after\ H
\end{equation}
where $F$ and $G$ are fluent formulas and $H$ is a formula, provided that $F$ does not contain statically determined constants and $H$ does not contain initpf constants.

A {\em pf constant declaration} is an expression of the form
\begin{equation}\label{eq:pf-declare-no-time}
   \caused\ \j{c}=\{v_1:p_1, \dots, v_n:p_n\}
\end{equation}
where $\j{c}$ is a pf constant with domain $\{v_1, \dots, v_n\}$, $0<p_i<1$ for each $i\in\{1, \dots, n\}$\footnote{We require $0<p_i<1$ for each $i\in\{1, \dots, n\}$ for the sake of simplicity. On the other hand, if $p_i=0$ or $p_i=1$ for some $i$, that means either $v_i$ can be removed from the domain of $c$ or there is not really a need to introduce $c$ as a pf constant. So this assumption does not really sacrifice expressivity.}, and $p_1+\cdots+p_n=1$. In other words, \eqref{eq:pf-declare-no-time} describes the probability distribution of $c$.

An {\em initpf constant declaration} is an expression of the form (\ref{eq:pf-declare-no-time}) where $c$ is an initpf constant. 

An {\em initial static law} is an expression of the form
\begin{equation}\label{eq:init-static-law}
\init\ F\ \iif\ G
\end{equation}
where $F$ is a fluent constant and $G$ is a formula that contains neither action constants nor pf constants. 

A {\em causal law} is a static law, a fluent dynamic law, a pf constant  declaration, an initpf constant declaration, or an initial static law. An {\em action description} is a finite set of causal laws.

We use $\sigma^{fl}$ to denote the set of fluent constants, $\sigma^{act}$ to denote the set of action constants, $\sigma^{pf}$ to denote the set of pf constants, and $\sigma^{initpf}$ to denote the set of initpf constants. For any signature $\sigma^\prime$ and any $i\in\{0, \dots, m\}$, we use $i:\sigma^\prime$ to denote the set
$\{i:a \mid a\in\sigma^\prime\}$.

By $i:F$ we denote the result of inserting $i:$ in front of every occurrence of every constant in formula $F$. This notation is straightforwardly extended when $F$ is a set of formulas.

\subsubsection{Semantics of $p\cal{BC}+$}
Given a non-negative integer $m$ denoting the maximum length of histories, the semantics of an action description $D$ in $p{\cal BC}$+ is defined by a reduction to multi-valued probabilistic program $Tr(D, m)$, which is the union of two subprograms $D_m$ and $D_{init}$ as defined below. 

For an action description $D$ of a signature $\sigma$, we define a sequence of multi-valued probabilistic program $D_0, D_1, \dots,$ so that the stable models of $D_m$ can be identified with the paths in the transition system described by $D$.

The signature $\sigma_m$ of $D_m$ consists of atoms of the form $i:c=v$ such that
\begin{itemize}
\item for each fluent constant $c$ of $D$, $i\in\{0, \dots, m\}$ and $v\in Dom(c)$,
\item for each action constant or pf constant $c$ of $D$, $i\in\{0, \dots, m-1\}$ and $v\in Dom(c)$.
\end{itemize}

For $x\in \{act, fl, pf\}$, we use $\sigma^{x}_m$ to denote the subset of $\sigma_m$
\[
\{i:c=v \mid \text{$i:c=v\in \sigma_m$ and $c\in\sigma^{x}$}\}.
\]
For $i\in\{0, \dots, m\}$, we use $i:\sigma^{x}$ to denote the subset of $\sigma_m^{x}$
\[
\{i:c=v\mid i:c=v\in \sigma_m^{x}\}.
\]

We define $D_m$ to be the multi-valued probabilistic program  $\langle PF, \Pi\rangle$, where $\Pi$ is the conjunction of
\begin{equation}\label{eq:static-law-asp}
i:F \leftarrow i:G
\end{equation}
for every static law (\ref{eq:static-law}) in $D$ and every $i\in\{0, \dots, m\}$,
\begin{equation}\label{eq:fluent-dynamic-law-asp}
i\!+\!1:F\leftarrow (i\!+\!1:G)\wedge(i:H)
\end{equation}
for every fluent dynamic law (\ref{eq:fluent-dynamic-law}) in $D$ and every $i\in\{0, \dots, m-1\}$,
\begin{equation}\label{eq:init-fluent-choice}
\{0\!:\!c=v\}^{\rm ch}
\end{equation}
for every regular fluent constant $c$ and every $v\in Dom(c)$,
\begin{equation}\label{eq:action-choice}
\{i:c=\true\}^{\rm ch}, \ \ \ \ 
\{i:c=\false\}^{\rm ch}
\end{equation}
for every action constant $c$, and $PF$ consists of 
\begin{equation}\label{eq:prod-declaration-mvpp}
   p_1::\ i:pf=v_1 \mid \dots \mid p_n::\ i:pf=v_{n}
\end{equation}
($i=0,\dots,m-1$) for each pf constant declaration \eqref{eq:pf-declare-no-time} in $D$ that describes the probability distribution of $\j{pf}$.

In addition, we define the program $D_{init}$, whose signature is $0\!:\!\sigma^{initpf}\cup 0\!:\!\sigma^{fl}$.
$D_{init}$ is the multi-valued probabilistic program
\[
D_{init} = \langle PF^{init}, \Pi^{init}\rangle
\]
where $\Pi^{init}$ consists of the rule
\[
\bot\leftarrow \neg(0\!:\!F)\land 0\!:\!G
\]
for each initial static law (\ref{eq:init-static-law}),
and $PF^{init}$ consists of 
\[
p_1::\ 0\!:\!pf=v_1\ \ \mid\ \  \dots\ \  \mid\ \  p_n::\ 0\!:\!pf=v_n
\]
for each initpf constant declaration (\ref{eq:pf-declare-no-time}).

We define $Tr(D, m)$ to be the union of the two multi-valued probabilistic program \\
$
\langle  PF\cup PF^{init}, \Pi\cup\Pi^{init} \rangle.
$

For any $\lpmln$ program $\Pi$ of signature $\sigma$ and a value assignment $I$ to a subset $\sigma'$ of $\sigma$, we say $I$ is a {\em residual (probabilistic) stable model} of $\Pi$ if there exists a value assignment $J$ to $\sigma\setminus \sigma'$ such that $I\cup J$ is a (probabilistic) stable model of $\Pi$.

For any value assignment $I$ to constants in $\sigma$, by $i\!:\!I$ we denote the value assignment to constants in $i\!:\!\sigma$ so that $i\!:\!I\models (i\!:\!c)=v$ iff $I\models c=v$.

We define a {\em state} as an interpretation $I^{fl}$ of $\sigma^{fl}$ such that $0\!:\!I^{fl}$ is a residual (probabilistic) stable model of $D_0$. A {\em transition} of $D$ is a triple $\langle s, e, s^\prime\rangle$  where $s$ and $s^\prime$ are interpretations of $\sigma^{fl}$ and $e$ is a an interpretation of $\sigma^{act}$ such that $0\!:\!s \cup 0\!:\!e \cup 1:s^\prime$ is a residual stable model of $D_1$. A {\em pf-transition} of $D$ is a pair $(\langle s, e, s^\prime\rangle, pf)$, where $pf$ is a value assignment to $\sigma^{pf}$ such that $0\!:\!s\cup 0\!:\!e \cup 1:s^\prime \cup 0\!:\!pf$ is a stable model of $D_1$. 

A {\em probabilistic transition system} $T(D)$ represented by a probabilistic action description $D$ is a labeled directed graph such that the vertices are the states of $D$, and the edges are obtained from the transitions of $D$: for every transition $\langle s, e, s^\prime\rangle$  of $D$, an edge labeled $e: p$ goes from $s$ to $s^\prime$, where $p=Pr_{D_1}(1\!:\!s^\prime \mid 0\!:\!s, 0\!:\!e)$. The number $p$ is called the {\em transition probability} of $\langle s, e ,s^\prime\rangle$ .

The soundness of the definition of a probabilistic transition system relies on the following proposition. 
\begin{prop}\label{prop:state-in-transition}
For any transition $\langle s, e, s^\prime \rangle$, $s$ and $s^\prime$ are states.
\end{prop}

We make the following simplifying assumptions on action descriptions:

\begin{enumerate}
\item {\bf No concurrent execution of actions}: For all transitions $\langle s, e, s'\rangle$, we have $e\models a\!=\!\true$ for at most one action constant $a$; 
%
\item {\bf Nondeterministic transitions are determined by pf constants}: For any state $s$, any value assignment $e$ of $\sigma^{act}$, and any value assignment $pf$ of $\sigma^{pf}$, there exists exactly one state $s^\prime$ such that $(\langle s, e, s^\prime\rangle, pf)$ is a pf-transition;
\item {\bf Nondeterminism on initial states are determined by initpf constants}: For any value assignment $pf_{init}$ of $\sigma^{initpf}$, there {exists exactly one value assignment $fl$ of $\sigma^{fl}$ such that $0\!:\!pf_{init}\cup 0\!:\!fl$ is a stable model of $D_{init}\cup D_0$.}
\end{enumerate}

For any state $s$, any value assignment $e$ of $\sigma^{act}$ such that at most one action is true, and any value assignment $pf$ of $\sigma^{pf}$, we use $\phi(s, e, pf)$ to denote the state $s'$ such that $(\langle s, a, s^\prime\rangle, pf)$ is a pf-transition (According to Assumption 2, such $s^\prime$ must be unique). For any interpretation $I$, $i\in \{0, \dots, m\}$ and any subset $\sigma^\prime$ of $\sigma$, we use $I|_{i:\sigma^\prime}$ to denote the value assignment of $I$ to atoms in $i:\sigma^\prime$. Given any value assignment $TC$ of $0\!:\!\sigma^{initpf}\cup \sigma^{pf}_m$and a value assignment $A$ of $\sigma_m^{act}$, we construct an interpretation $I_{TC\cup A}$ of $Tr(D, m)$ that satisfies $TC \cup A$ as follows:
\begin{itemize}
\item  For all atoms $p$ in $\sigma^{pf}_m\cup 0\!:\!\sigma^{initpf}$, 
           we have $I_{TC\cup A}(p) = TC(p)$;
\item  For all atoms $p$ in $\sigma_m^{act}$, we have $I_{TC\cup A}(p) = A(p)$;
\item $(I_{TC\cup A})|_{0:\sigma^{fl}}$ is the assignment such that $(I_{TC\cup A})|_{0:\sigma^{fl}\cup 0:\sigma^{initpf}}$ is a stable model of $D_{init}\cup D_0$.
\item For each $i\in \{1, \dots, m\}$, $$(I_{TC\cup A})|_{i:\sigma^{fl}} = \phi((I_{TC\cup A})|_{(i-1):\sigma^{fl}}, (I_{TC\cup A})|_{(i-1):\sigma^{act}}, (I_{TC\cup A})|_{(i-1):\sigma^{pf}}).$$
\end{itemize}
By Assumptions 2 and 3, the above construction produces a unique interpretation. 

It can be seen that in the multi-valued probabilistic program $Tr(D, m)$ translated from $D$, the probabilistic constants  are $0\!:\!\sigma^{initpf}\cup \sigma^{pf}_m$. We thus call the value assignment of an interpretation $I$ on $0\!:\!\sigma^{initpf}\cup \sigma^{pf}_m$ the {\em total choice} of $I$. The following theorem asserts that the probability of a stable model under $Tr(D, m)$ can be computed by simply dividing the probability of the total choice associated with the stable model by the number of choice of actions.

\begin{thm}\label{thm:path-probability}
For any value assignment $TC$ of $ 0\!:\!\sigma^{initpf}\cup\sigma^{pf}_m$ and any value assignment $A$ of $\sigma_m^{act}$, there exists exactly one stable model $I_{TC\cup A}$ of $Tr(D, m)$ that satisfies $TC\cup A$, and the probability of $I_{TC\cup A}$ is
\[
Pr_{Tr(D, m)}(I_{TC\cup A}) = \frac{\underset{c=v\in TC}{\prod}M(c=v)}{(|\sigma^{act}| + 1)^{m}}.
\]
\end{thm}

The following theorem tells us that the conditional probability of transiting from a state $s$ to another state $s^\prime$ with action $e$ remains the same for all timesteps, i.e., the conditional probability of $i\!+\!1\!:\!s^\prime$ given $i:s$ and $i:e$ correctly represents the transition probability from $s$ to $s^\prime$ via $e$ in the transition system.

\begin{thm}\label{thm:transition-probability}
For any state $s$ and $s^\prime$, and action $e$, we have
\[
Pr_{Tr(D, m)}(i\!+\!1\!:\!s^\prime\mid i:s, i:e) = Pr_{Tr(D, m)}(j\!+\!1\!:\!s^\prime\mid j:s, j:e)
\]
for any $i, j\in\{0, \dots, m-1\}$ such that $Pr_{Tr(D, m)}(i:s)> 0$ and $Pr_{Tr(D, m)}(j:s)> 0$.
\end{thm}

For every subset $X_m$ of $\sigma_m\setminus\sigma^{pf}_m$, let $X^i(i < m)$ be the triple consisting of
\begin{itemize}
\item the set consisting of atoms $A$ such that $i:A$ belongs to $X_m$ and $A\in \sigma^{fl}$;
\item the set consisting of atoms $A$ such that $i:A$ belongs to $X_m$ and $A\in \sigma^{act}$;
\item the set consisting of atoms $A$ such that $i\!+\!1\!:\!A$ belongs to $X_m$ and $A\in \sigma^{fl}$.
\end{itemize}
Let $p(X^i)$ be the transition probability of $X^i$, $s_0$ is the interpretation of $\sigma^{fl}_0$ defined by $X^0$, and $e_i$ be the interpretations of $i:\sigma^{act}$ defined by $X^{i}$.

Since the transition probability remains the same, the probability of a path given a sequence of actions can be computed from the probabilities of transitions.

\begin{cor}\label{thm:reduce2transition}
For every $m\geq 1$, $X_m$ is a residual (probabilistic) stable model of $Tr(D, m)$ iff $X^0, \dots, X^{m-1}$ are transitions of $D$ and $0\!:\!s_0$ is a residual stable model of $D_{init}$. Furthermore, 
\[
Pr_{Tr(D, m)}(X_m\mid 0\!:\!e_0, \dots, m-1\!:\!e_{m-1}) = p(X^0)\times\dots\times p(X^m)\times Pr_{Tr(D, m)}(0\!:\!s_0).
\]
\end{cor}

\subsubsection{$\pbcp$ with Utility}

Wang and Lee \citeyear{wang19elaboration} has extended $\pbcp$ with the notion of utility as follows.

We extend $\pbcp$ by introducing the following expression called {\em utility law} that assigns a reward to transitions:
\begin{equation}\label{eq:utility-law}
\reward\ v\ {\bf if}\ F\ {\bf after}\ G
\end{equation}
where $v$ is a real number representing the reward, $F$ is a formula that contains fluent constants only, and $G$ is a formula that contains fluent constants and action constants only (no pf, no initpf constants). We extend the signature of $Tr(D, m)$ with a set of atoms of the form~\eqref{eq:utility-atoms}. 
We turn a utility law of the form \eqref{eq:utility-law} into the $\lpmln$ rule
\begin{equation}\label{eq:utility-law-lpmln}
\alpha: {\tt utility}(v, i+1, id)\ \ar\ (i+1:F)\wedge(i:G)
\end{equation}
where $id$ is a unique number assigned to the $\lpmln$ rule and $i\in\{0,\dots, m\!-\!1\}$.

Given a nonnegative integer $m$ denoting the maximum timestamp, a $\pbcp$ action description $D$ with utility over multi-valued propositional signature $\sigma$ is defined as a high-level representation of the $\dtlpmln$ program $(Tr(D, m), \sigma^{act}_m)$.

%

We extend the definition of a probabilistic transition system as follows: A {\em probabilistic transition system} $T(D)$ represented by a probabilistic action description $D$ is a labeled directed graph such that the vertices are the states of $D$, and the edges are obtained from the transitions of $D$: for every transition $\langle s, e, s^\prime\rangle$  of $D$, an edge labeled $e: p, u$ goes from $s$ to $s^\prime$, where $p=Pr_{D_1}(1\!:\!s^\prime \mid 0\!:\!s\wedge 0\!:\!e)$ and $u=E[U_{D_1}(0\!:\!s\wedge 0\!:\!e\wedge 1\!:\!s')]$. The number $p$ is called the {\em transition probability} of $\langle s, e ,s^\prime\rangle$, denoted by $p(s, e ,s^\prime)$, and the number $u$ is called the {\em transition reward} of $\langle s, e ,s^\prime\rangle$, denoted by $u(s, e ,s^\prime)$.

\section{{\sc pbcplus2pomdp} in Compositional Way} \label{sec:compo}

In particular, the inputs of {\sc pbcplus2pomdp(compo)} include the following: 
\begin{itemize}
    \item $\lpmln$ program $\Pi(m)$, parameterized with maximum timestep $m$, that contains $\lpmln$ translation of fluent dynamic laws, observation dynamic laws and utility laws with no occurrence of action constant, and static laws, as well as pf constant declarations of pf constants that occur in those causal laws (see Figure~\ref{fig:pbcplus-causal-laws});
    \item For each group of actions $a_i\in$ in $a_1$,$\dots$, $a_n$, an $\lpmln$ program $\Pi_i(m)\cup C_i(m)$, parameterized with maximum timestep $m$; $\Pi_i(m)$ contains translation of fluent dynamic laws, observation dynamic laws and utility laws where only actions in $a_i$ can occur in the body, as well as pf constant declarations of pf constants that occurs in those causal laws; $C_i(m)$ contains choice rules (possibly with cardinality bounds) to generate exactly one action in the group $a_i$; It is up to the user how to group the actions; 
    \item Discount factor.
\end{itemize}
The system outputs the POMDP definition $M(D)$, so that $D_m=\Pi(m)\cup \Pi_1(m)\cup\dots\Pi_n(m)\cup C(m)$, where $C(m)$ is the choice rule with cardinality constraint to generate at most one action in $a_1, \dots, a_n$ for each timestep $i\in\{0, \dots, m-1\}$. The transition probabilities, observation probabilities and reward function of $M(D)$ are obtained by conjoining those from each of $\Pi\cup \Pi_i\cup C_i$ ($i\in\{1, \dots, n\}$). 

Formally, let ${\bf S}$, $\Omega$, $P_{M(D)}$, $O_{M(D)}$, $R_{M(D)}$ be the set of states, the set of observations, transition probabilities, observation probabilities and reward function of $M(D)$, resp. system {\sc pbcplus2pomdp} calls {\sc lpmln2asp} first to solve $\Pi(0)$ to obtain ${\bf S}$, and then $\Pi(1)\cup\Pi_i(1)\cup C_i(1)$ to obtain $P_{M(D)}$, $O_{M(D)}$, $R_{M(D)}$ as follows:
\[
P_{M(D)}(s, a, s') = P_{\Pi(1)\cup \Pi_i(1)\cup C_i(1)}(1:s' \mid 0:s, 0:a)
\]
\[
O_{M(D)}(s, a, o) = P_{\Pi(1)\cup \Pi_i(1)\cup C_i(1)}(1:o\mid 1:s, 0:a)
\]
\[
R_{M(D)}(s, a, s') = E[U_{\Pi(1)\cup \Pi_i(1)\cup C_i(1)}(0:s, 0:a, 1:s')]
\]
for each $a\in a_i$, $s, s'\in {\bf S}$ and $o\in\Omega$.


\begin{example}
For the dialog example, we group the actions as follows:
   $\{\j{ConfirmItem}(i)\mid i\in \j{Item}\}$,
   $\{\j{ConfirmPerson}(p)\mid p\in \j{Person}\}$,
   $\{\j{ConfirmRoom}(r)\mid r\in \j{Room}\}$,
   $\{\j{WhichItem}\}$,
   $\{\j{WhichPerson}\}$,
   $\{\j{WhichRoom}\}$,
   $\{\j{Deliver}(i, p, r)\mid i\in \j{Item}, p\in \j{Person}, r\in \j{Room}\}$.

$\Pi$ is
\begin{lstlisting}
astep(0..m-1).
step(0..m).
boolean(t; f).
item(coffee; coke; cookies; burger).
person(alice; bob; carol).
room(r1; r2; r3).

% UEC
:- obs_Item(X1, I), obs_Item(X2, I), X1 != X2.
:- not obs_Item(coffee, I), not obs_Item(coke, I),
   not obs_Item(cookies, I), not obs_Item(burger, I),
   not obs_Item(na, I), step(I).
:- obs_Person(X1, I), obs_Person(X2, I), X1 != X2.
:- not obs_Person(alice, I), not obs_Person(bob, I),
   not obs_Person(carol, I), not obs_Person(na, I),
   step(I).
:- obs_Room(X1, I), obs_Room(X2, I), X1 != X2.
:- not obs_Room(r1, I), not obs_Room(r2, I),
   not obs_Room(r3, I), not obs_Room(na, I),
   step(I).
:- obs_Confirmed(X1, I), obs_Confirmed(X2, I), X1 != X2.
:- not obs_Confirmed(yes, I), not obs_Confirmed(no, I),
   not obs_Confirmed(na, I), step(I).

:- fl_ItemReq(X1, I), fl_ItemReq(X2, I), X1 != X2.
:- not fl_ItemReq(coffee, I), not fl_ItemReq(coke, I),
   not fl_ItemReq(cookies, I), not fl_ItemReq(burger, I),
   not fl_ItemReq(na, I), step(I).
:- fl_PersonReq(X1, I), fl_PersonReq(X2, I), X1 != X2.
:- not fl_PersonReq(alice, I), not fl_PersonReq(bob, I),
   not fl_PersonReq(carol, I), not fl_PersonReq(na, I),
   step(I).
:- fl_RoomReq(X1, I), fl_RoomReq(X2, I), X1 != X2.
:- not fl_RoomReq(r1, I), not fl_RoomReq(r2, I),
   not fl_RoomReq(r3, I), not fl_RoomReq(na, I),
   step(I).
:- fl_Terminated(X1, I), fl_Terminated(X2, I), X1 != X2.
:- not fl_Terminated(t, I), not fl_Terminated(f, I), step(I).

%% No two observations can occur at the same time step
:- obs_Item(It, I), obs_Person(P, I), It != na, P != na.
:- obs_Item(It, I), obs_Room(R, I), It != na, R != na.
:- obs_Item(It, I), obs_Confirmed(C, I), It != na, C != na.
:- obs_Person(P, I), obs_Room(R, I), P != na, R != na.
:- obs_Person(P, I), obs_Confirmed(C, I), P != na, C != na.
:- obs_Room(R, I), obs_Confirmed(C, I), R != na, C != na.

% Inertial Fluents
{fl_ItemReq(It, I+1)} :- fl_ItemReq(It, I), astep(I).
{fl_PersonReq(P, I+1)} :- fl_PersonReq(P, I), astep(I).
{fl_RoomReq(R, I+1)} :- fl_RoomReq(R, I), astep(I).
{fl_Terminated(B, I+1)} :- fl_Terminated(B, I), astep(I).

% Initial value of regular fluents and observation constants are exogenous
{fl_Terminated(B, 0)} :- boolean(B).
{fl_ItemReq(It, 0)} :- item(It).
{fl_PersonReq(P, 0)} :- person(P).
{fl_RoomReq(R, 0)} :- room(R).
{obs_Item(It, 0)} :- item(It).
{obs_Person(P, 0)} :- person(P).
{obs_Room(R, 0)} :- room(R).
{obs_Confirmed(yes, 0); obs_Confirmed(no, 0)}.

% By default, observation constant has na value
{obs_Item(na, I)} :- step(I).
{obs_Person(na, I)} :- step(I).
{obs_Room(na, I)} :- step(I).
{obs_Confirmed(na, I)} :- step(I).
\end{lstlisting}

$\Pi_1$ contains definition of action $Ask2ConfirmItem$:
\begin{lstlisting}
% Action: ConfirmItem
:- c(It, X1, I), act_ConfirmItem(It, X2, I), X1 != X2.
:- not act_ConfirmItem(It, t, I), not act_ConfirmItem(It, f, I), item(It), astep(I).

:- pf_ConfirmWhenCorrect(X1, I), pf_ConfirmWhenCorrect(X2, I), X1 != X2.
:- not pf_ConfirmWhenCorrect(yes, I), not pf_ConfirmWhenCorrect(no, I), astep(I).
:- pf_ConfirmWhenIncorrect(X1, I), pf_ConfirmWhenIncorrect(X2, I), X1 != X2.
:- not pf_ConfirmWhenIncorrect(yes, I), not pf_ConfirmWhenIncorrect(no, I), astep(I).

@log(0.8) pf_ConfirmWhenCorrect(yes, I) :- astep(I).
@log(0.2) pf_ConfirmWhenCorrect(no, I) :- astep(I).

@log(0.2) pf_ConfirmWhenIncorrect(yes, I) :- astep(I).
@log(0.8) pf_ConfirmWhenIncorrect(no, I) :- astep(I).

obs_Confirmed(C, I+1) :- fl_ItemReq(It, I+1), fl_Terminated(f, I+1),
           act_ConfirmItem(It, t, I), pf_ConfirmWhenCorrect(C, I).
obs_Confirmed(C, I+1) :- fl_ItemReq(It, I+1), fl_Terminated(f, I+1),
           act_ConfirmItem(It1, t, I), It1 != It, pf_ConfirmWhenIncorrect(C, I).

{act_ConfirmItem(It, B, I)} :- item(It), boolean(B), astep(I).
:- not 1{act_ConfirmItem(It, t, I) : item(It)}1, astep(I).
\end{lstlisting}

Similarly, $\Pi_2$, $\Pi_3$ contains definition of actions $ConfirmPerson$ and $ConfirmRoom$.

$\Pi_4$ contains definition of actions $WhichItem(t)$:
\begin{lstlisting}
% Action WhichItem
:- act_WhichItem(X1, I), act_WhichItem(X2, I), X1 != X2.
:- not act_WhichItem(t, I), not act_WhichItem(f, I), astep(I).

:- pf_WhichItem(It, X1, I), pf_WhichItem(It, X2, I), X1 != X2.
:- not pf_WhichItem(It, coffee, I), not pf_WhichItem(It, coke, I),
   not pf_WhichItem(It, cookies, I), not pf_WhichItem(It, burger, I),
   item(It), astep(I).

@log(0.7) pf_WhichItem(coffee, coffee, I) :- astep(I).
@log(0.1) pf_WhichItem(coffee, coke, I) :- astep(I).
@log(0.1) pf_WhichItem(coffee, cookies, I) :- astep(I).
@log(0.1) pf_WhichItem(coffee, burger, I) :- astep(I).
@log(0.1) pf_WhichItem(coke, coffee, I) :- astep(I).
@log(0.7) pf_WhichItem(coke, coke, I) :- astep(I).
@log(0.1) pf_WhichItem(coke, cookies, I) :- astep(I).
@log(0.1) pf_WhichItem(coke, burger, I) :- astep(I).
@log(0.1) pf_WhichItem(cookies, coffee, I) :- astep(I).
@log(0.1) pf_WhichItem(cookies, coke, I) :- astep(I).
@log(0.7) pf_WhichItem(cookies, cookies, I) :- astep(I).
@log(0.1) pf_WhichItem(cookies, burger, I) :- astep(I).
@log(0.1) pf_WhichItem(burger, coffee, I) :- astep(I).
@log(0.1) pf_WhichItem(burger, coke, I) :- astep(I).
@log(0.1) pf_WhichItem(burger, cookies, I) :- astep(I).
@log(0.7) pf_WhichItem(burger, burger, I) :- astep(I).

obs_Item(It1, I+1) :- fl_ItemReq(It, I+1), fl_Terminated(f, I+1), 
           act_WhichItem(t, I), pf_WhichItem(It, It1, I).

%{act_WhichItem(B, I)} :- boolean(B), astep(I).
act_WhichItem(t, I) :- astep(I).
\end{lstlisting}
Similarly, $\Pi_5$ and $\Pi_6$ contain definitions of actions $WhichPerson(t)$ and $WhichRoom(t)$.

$\Pi_7$ contains definitions of action $Deliver(i, p, r)$:
\begin{lstlisting}
% Action: Deliver
:- act_Deliver(It, P, R, X1, I), act_Deliver(It, P, R, X2, I), X1 != X2.
:- not act_Deliver(It, P, R, t, I), not act_Deliver(It, P, R, f, I), item(It), person(P), room(R), astep(I
).

utility(1, I+1, It) :-  fl_ItemReq(It, I+1), act_Deliver(It, P, R, t, I), fl_Terminated(f, I).
utility(1, I+1, P) :-  fl_PersonReq(P, I+1), act_Deliver(It, P, R, t, I), fl_Terminated(f, I).
utility(1, I+1, R) :-  fl_RoomReq(R, I+1), act_Deliver(It, P, R, t, I), fl_Terminated(f, I).

fl_Terminated(t, I+1) :- act_Deliver(It, P, R, t, I).

{act_Deliver(It, P, R, B, I)} :- item(It), person(P), room(R), boolean(B), astep(I).
:- not 1{act_Deliver(It, P, R, t, I) : item(It), person(P), room(R)}1, astep(I).
\end{lstlisting}

$\Pi_8$ contains definitions of no-action:
\begin{lstlisting}
act_noact(I) :- astep(I).
\end{lstlisting}
With this way of grouping actions, system {\sc pbcplus2pomdp(compo)} can generate POMDP for this example with $\sim 5$ minutes.
\end{example}

\section{Tiger Example}

\begin{example}\label{eg:2-tiger}
({\bf Two Tigers Example}). Consider a variant of the well-known tiger example extended with two tigers. Each of the three doors has either a tiger or a prize behind. 
The agent can open either of the three doors. The agent can also listen to get a better idea of where the tiger is. Listening yields the correct information about where each of the two tigers is with probability $0.85$. 
This example can be represented in the extended $p\cal{BC}+$ as follows:

\vspace{0.2cm}
\hrule
\begin{tabbing}
Notation:  $l, l_1, l_2, l_3$ range over ${\tt Left}$, ${\tt Middle}$, ${\tt Right}$, $y$ ranges over ${\tt Tiger1}$, ${\tt Tiger2}$\\
Observation constants:         \hskip 4cm  \=Domains:\\
$\;\;\;$ $\j{TigerPositionObserved}(y)$                 \>$\;\;\;$ $\{{\tt Left}, {\tt Middle}, {\tt Right}, {\tt NA}\}$\\ 
Regular fluent constants:          \hskip 4cm  \=Domains:\\
$\;\;\;$ $\j{TigerPosition}(y)$                 \>$\;\;\;$ $\{{\tt Left}, {\tt Middle}, {\tt Right}\}$\\ 
Action constants:                          \>Domains:\\
$\;\;\;$ $\j{Listen}$  \>$\;\;\;$ Boolean\\
$\;\;\;$ $\j{OpenDoor}(l)$  \>$\;\;\;$ Boolean\\ 
Pf constants:                          \>Domains:\\
$\;\;\;$ $\j{Pf\_Listen}$                    \>$\;\;\;$ Boolean\\
$\;\;\;$ $\j{Pf\_FailedListen}(y)$                    \>$\;\;\;$ $\{{\tt Left}, {\tt Middle}, {\tt Right}\}$
\end{tabbing}
\hrule
\vspace{0.2cm}

A reward of $10$ is obtained for opening the door with no tiger behind.
\begin{align}
\nonumber &{\reward}\ 10\ \iif\ \j{TigerPosition}({\tt Tiger1})\mvis l_1\land \j{TigerPosition}({\tt Tiger2})\mvis l_2\ \after\ \j{OpenDoor}(l_3)\\
\nonumber & \ \ \ \ \ \ \ (l_1 \neq l_3,  l_2\neq l_3).
\end{align}
A penalty of $100$ is imposed on opening a door with a tiger behind.
\begin{align}
\nonumber &{\reward}\ -\!\!100\ \iif\ \j{TigerPosition}(y)\mvis l\ \after\ \j{OpenDoor}(l).
\end{align}
Executing the action $\j{Listen}$ has a small penalty of $1$.
\begin{align}
\nonumber &{\reward}\ -\!\!1\ \iif\ \top\ \after\ \j{Listen}.
\end{align}
Two tigers cannot be in the same position.
\begin{align}
\nonumber &{\caused}\ \bot\ \iif\ \j{TigerPosition}({\tt Tiger1})\mvis l\land \j{TigerPosition}({\tt Tiger2})\mvis l.
\end{align}
Successful listening reveals the positions of the two tigers.
\begin{align}
\nonumber &{\observed}\ \j{TigerPositionObserved}(y)\mvis l\ {\bf if}\ \j{TigerPosition}(y)\mvis l\ {\bf after}\ \j{Listen}\land \j{Pf\_Listen}.
\end{align}
Failed listening yields a random position for each tiger.
\begin{align}
\nonumber &\caused\ \j{Pf\_FailedListen}(y)=\{{\tt Left}: \frac{1}{3}, {\tt Middke}: \frac{1}{3}, {\tt Right}: \frac{1}{3}\},\\
\nonumber &{\observed}\ \j{TigerPositionObserved}(y)\mvis l\ {\bf if}\ \top\ {\bf after}\ \j{Listen}\land \sim\j{Pf\_Listen}\land \j{Pf\_FailedListen(y)} \mvis  l.
\end{align}
The positions of tigers observe the commonsense law of inertia.
\begin{align}
\nonumber &\inertial\ \j{TigerPosition}(y).
\end{align}
The action $\j{Listen}$ has a success rate of $0.85$.
\begin{align}
\nonumber &\caused\ \j{Pf\_Listen}=\{\true: 0.85, \false: 0.15\}.
\end{align}

\end{example}

\end{appendix}

\end{document}